\documentclass[11pt]{article}
\usepackage[top=1in, bottom=1in, left=1in, right=1in]{geometry}
\usepackage[numbers]{natbib}
\usepackage{amsmath}
\usepackage{amssymb}
\usepackage{mathtools}
\usepackage{amsthm}
\usepackage{parskip}

\usepackage{color-edits}
\addauthor[Anay]{am}{blue}

\usepackage{hyperref}
\usepackage[capitalize,noabbrev,sort]{cleveref}

\usepackage{algorithmic}
\usepackage{mathpazo}

\theoremstyle{plain}
\newtheorem{theorem}{Theorem}[section]

\newtheorem{lemma}[theorem]{Lemma}

\theoremstyle{definition}
\newtheorem{definition}[theorem]{Definition}
\newtheorem{assumption}[theorem]{Assumption}
\theoremstyle{remark}
\newtheorem{remark}[theorem]{Remark}

\usepackage[textsize=tiny]{todonotes}

\usepackage[utf8]{inputenc} %
\usepackage[T1]{fontenc}    %
\usepackage{hyperref}       %
\usepackage{url}            %
\usepackage{booktabs}       %
\usepackage{amsfonts}       %
\usepackage{nicefrac}       %
\usepackage{microtype}      %
\usepackage{xcolor}         %

\usepackage{savesym}
\savesymbol{Bbbk}
\usepackage{booktabs}
\usepackage{microtype}
\usepackage{multicol}
\usepackage{multirow}
\usepackage{tcolorbox}
\usepackage{savesym}

\usepackage{amsmath}
\usepackage{amssymb}
\usepackage{mathtools}
\usepackage{amsthm}

\usepackage{thmtools}
\usepackage{thm-restate}

\usepackage{multicol}
\usepackage{multirow}
\usepackage{tcolorbox}

\usepackage{algorithm} 

\usepackage{amsfonts}
\usepackage{url}
\usepackage{tikz}
\usepackage{mathrsfs}
\usepackage{nicefrac,bbm}
\usepackage{pifont,enumitem}
\usepackage[framemethod=TikZ]{mdframed}
\usepackage{color,fancybox}

\usepackage[normalem]{ulem}

\usepackage{xurl}
\usepackage{scalerel}
\usetikzlibrary{math}
   
\usepackage{bbding}

\usepackage{array} 
\usepackage{dsfont}

\newcommand{\yesnum}{\addtocounter{equation}{1}\tag{\theequation}}
\newcommand{\tagnum}[1]{\addtocounter{equation}{1}{\tag{#1)\ \ (\theequation}}}
\makeatletter
\newcommand{\customlabel}[2]{%
\protected@write \@auxout {}{\string \newlabel {#1}{{#2}{\thepage}{#2}{#1}{}} }%
\hypertarget{#1}{}
}
\makeatother

\mdfdefinestyle{FrameBox}{ linecolor=white, outerlinewidth=0pt, roundcorner=5pt,
innertopmargin=5pt, innerbottommargin=5pt,
innerrightmargin=10pt, innerleftmargin=10pt, backgroundcolor=white}

\mdfdefinestyle{FrameBox2}{ linecolor=black, outerlinewidth=0pt, roundcorner=5pt,
innertopmargin=5pt, innerbottommargin=5pt,
innerrightmargin=10pt, innerleftmargin=10pt, backgroundcolor=white}

\mdfdefinestyle{FrameQuestion2}{linecolor=white, outerlinewidth=0pt,
roundcorner=5pt,
innertopmargin=5pt, innerbottommargin=0pt,
innerrightmargin=16pt, innerleftmargin=16pt, backgroundcolor=white}

\newcommand{\white}[1]{\textcolor{white}{#1}}

\colorlet{RED}{red}
\colorlet{BLUE}{blue}

\newcommand{\N}{\mathbb{N}}

\newcommand{\R}{\mathbb{R}}

\newcommand{\cB}{\mathcal{B}}

\newcommand{\cD}{\mathcal{D}}

\newcommand{\cF}{\mathcal{F}}

\newcommand{\cX}{\mathcal{X}}
\newcommand{\cY}{\mathcal{Y}}

\newcommand{\evE}{\ensuremath{\mathscr{E}}}
\newcommand{\evF}{\ensuremath{\mathscr{F}}}
\newcommand{\evG}{\ensuremath{\mathscr{G}}}
\newcommand{\evH}{\ensuremath{\mathscr{H}}}

\newcommand{\nfrac}{\nicefrac}
\newcommand{\sfrac}[2]{#1/#2}
\newcommand{\st}{\mathrm{s.t.}}

\newcommand{\eps}{\varepsilon}
\renewcommand{\epsilon}{\varepsilon}

\newcommand{\argmax}{\operatornamewithlimits{arg\;max}}
\newcommand{\Ex}{\operatornamewithlimits{\mathbb{E}}}

\newcommand{\poly}{\mathop{\mbox{\rm poly}}}

\def\abs#1{\left| #1 \right|}
\def\sabs#1{| #1 |}

\newcommand{\sinparen}[1]{(#1)}
\newcommand{\sinbrace}[1]{\{#1\}}

\newcommand{\inparen}[1]{\left(#1\right)}
\newcommand{\inbrace}[1]{\left\{#1\right\}}
\newcommand{\insquare}[1]{\left[#1\right]}

\newcommand{\specialcell}[2][c]{%
  \begin{tabular}[#1]{@{}c@{}}#2\end{tabular}}

\newcommand{\zo}{\{0,1\}}

\newcommand{\wh}[1]{\widehat{#1}}

\newcommand{\hy}{\widehat{y}}

\newcommand{\prog}[1]{Program~\eqref{#1}}

\newcommand{\Comment}[1]{\textcolor{gray}{$\triangleright$#1}}

\newcommand{\Stackrel}[2]{\stackrel{\mathmakebox[\widthof{\ensuremath{#2}}]{#1}}{#2}}

\newif\ifconf
\conffalse

\ifconf

\else

\fi

\ifconf

\else

\fi

\newcommand{\eat}[1]{}

\crefname{section}{Section}{Sections}
\crefname{theorem}{Theorem}{Theorems}
\crefname{assumption}{Assumption}{Assumptions}
\crefname{lemma}{Lemma}{Lemmas}
\crefname{definition}{Definition}{Definitions}
\crefname{conjecture}{Conjecture}{Conjectures}
\crefname{corollary}{Corollary}{Corollaries}
\crefname{construction}{Construction}{Constructions}
\crefname{claim}{Claim}{Claims}
\crefname{observation}{Observation}{Observations}
\crefname{proposition}{Proposition}{Propositions}
\crefname{fact}{Fact}{Facts}
\crefname{question}{Question}{Questions}
\crefname{problem}{Problem}{Problems}
\crefname{remark}{Remark}{Remarks}
\crefname{example}{Example}{Examples}
\crefname{equation}{Equation}{Equations}
\crefname{appendix}{Appendix}{Appendices}
\crefname{algorithm}{Algorithm}{Algorithms}
\crefname{model}{Model}{Models}
\crefname{figure}{Figure}{Figures}  

\AfterEndEnvironment{definition}{\noindent\ignorespaces}
\AfterEndEnvironment{assumption}{\noindent\ignorespaces}
\AfterEndEnvironment{lemma}{\noindent\ignorespaces}
\AfterEndEnvironment{theorem}{\noindent\ignorespaces}
\AfterEndEnvironment{proposition}{\noindent\ignorespaces}
\AfterEndEnvironment{fact}{\noindent\ignorespaces}
\AfterEndEnvironment{question}{\noindent\ignorespaces}
\AfterEndEnvironment{proof}{\noindent\ignorespaces}

\newcommand{\calO}{\mathcal{O}}

\makeatletter
\def\moverlay{\mathpalette\mov@rlay}
\def\mov@rlay#1#2{\leavevmode\vtop{%
   \baselineskip\z@skip \lineskiplimit-\maxdimen
   \ialign{\hfil$\m@th#1##$\hfil\cr#2\crcr}}}
\newcommand{\charfusion}[3][\mathord]{
    #1{\IFx#1\mathop\vphantom{#2}\fi
        \mathpalette\mov@rlay{#2\cr#3}
      }
    \IFx#1\mathop\expandafter\displaylimits\fi}
\makeatother

\newcommand{\util}{\ensuremath{\text{Util}}}

\newcommand{\optOffline}{\ensuremath{f_{{\rm opt}}}}

\newlength\myindent
\newcommand{\expl}{\textsc{Exploit}}
\newcommand{\expr}{\textsc{Explore}}

\title{\bf Fair Classification with Partial Feedback:\\ An Exploration-Based Data Collection Approach}

\begin{document}

\author{
        \textbf{Vijay Keswani}\\[4mm] {Duke University}
        \and \hspace{-3mm}
        \textbf{Anay Mehrotra}\\[4mm] \hspace{-3mm} {Yale University}
        \and 
        \textbf{L. Elisa Celis}\\[4mm] {Yale University}
}

\date{}

\maketitle

\begin{abstract}

    In many predictive contexts (e.g., credit lending), true outcomes are \textit{only} observed for samples that were positively classified in the past. These past observations, in turn, form training datasets for classifiers that make future predictions. However, such training datasets lack information about the outcomes of samples that were (incorrectly) negatively classified in the past and can lead to erroneous classifiers. We present an approach that trains a classifier using available data and comes with a family of exploration strategies to collect outcome data about subpopulations that otherwise would have been ignored. For any exploration strategy, the approach comes with guarantees that (1) all sub-populations are explored, (2) the fraction of false positives is bounded, and (3) the trained classifier converges to a ``desired'' classifier. The right exploration strategy is context-dependent;  it can be chosen to improve learning guarantees and encode context-specific group fairness properties. Evaluation on real-world datasets shows that this approach consistently boosts the \textit{quality} of collected outcome data and improves the fraction of true positives for all groups, with only a small reduction in predictive utility.
\end{abstract}

\maketitle

\section{Introduction}\label{sec:intro} 
    Machine Learning (ML) classifiers are increasingly being used to aid decision-making in high-stake contexts such as credit lending, healthcare, and criminal justice.
    However, their real-world deployment still faces several practical challenges, including selective availability of ground truth labels, errors in collected data, and distribution or covariate shifts \cite{kleinberg2018human,fairmlbook,Nithya2021ModelWork}. 
    Of specific interest is the challenge of learning an accurate classifier in the \textit{partial feedback setting} where ground truth or outcome labels are \textit{only} observed for positively classified samples.
    For instance, a bank observes whether an individual defaults on loan payments or not only after granting them a loan, or a doctor observes the effect of a health intervention only if it is used.
    In these contexts, outcome labels are only observed for individuals who were ``positively classified'' in the past.

    Classification models are trained using prior outcome-labeled data, often under the assumption that future samples follow the same distribution as the training data. 
    However, in partial feedback settings, the unavailability of outcomes for unobserved samples can distort the data distribution 
    leading to large prediction errors.
    For example, suppose at year $j$, a set of applicants $S_j$ apply for a loan from a bank. 
    A classifier (trained on past data of loan applicants) assesses the default risk of all applicants and accepts the 
    applications $L_j \subset S_j$ and rejects 
    $U_j \coloneqq S_j {\setminus} L_j$.
    Following these decisions, the true default outcomes 
    are only observed for $L_j$, which is then
    added to training data to learn future classifiers.
    However, since $L_j$ can have a different distribution than $S_j$, the training data composed of $\{L_j\}_{j \in 1, 2, \dots}$ can misrepresent the population distribution and train erroneous classifiers.
    Classification errors due to partial feedback are indeed quite prevalent in practice.
    \citet{kleinberg2018human} and \citet{lakkaraju2017selective} note the difficulty of assessing bail decisions when true outcomes are only observed for released defendants.
    As described above, similar issues arise in the lending settings \cite{pacchiano2021neural}.
    In predictive policing, partial feedback leads to feedback loops that reinforce socioeconomic inequalities
    \cite{ensign2017runaway, marda2020data}.

    Issues arising from partial feedback are further compounded when decision-making processes are biased and display disparate performance across protected attributes (e.g., gender or race) \cite{mehrabi2021survey}.
    Biases affect many applications where ML algorithms are currently employed, e.g. bail decisions \cite{arnold2018racial}, loan applications \cite{denied2021markup}, and policing \cite{brantingham2017logic}.
    The impact of social biases can be significant in the partial feedback setting:
    Revisiting the lending example, suppose a classifier $f$ is used to predict the default risk for applicants $S_j$. Further, suppose that $f$ assigns disparately higher default risk to individuals from group $z_1$ compared to
    group $z_2$.
    As a result, a relatively larger fraction of individuals from $z_2$ will be assigned a loan, and since we only observe default outcomes for positively classified samples, we will have more information about the risk associated with individuals from $z_2$ than $z_1$.
    Future classifiers trained using this labeled data will propagate, \mbox{or even exacerbate, biases against $z_1$.}

    The pervasiveness of partial feedback in practice
    makes it necessary to study methods for \textit{fair} data collection and training in this setting. 
    One way of addressing this problem is by assigning positive predictions
    to all samples for which outcomes have not been observed in the past
    so as to explicitly observe their true outcomes.
    This way one can improve the quality of collected labeled data
    and, by extension, the predictive accuracy of trained classifiers.
    However, this approach can be highly impractical;
    positive predictions often entail high-stakes decisions (e.g., giving loans or predicting disease occurrence) and \textit{false positives}
    can have a significant negative impact on individual and institutional utility. %
    Hence, data collection in the partial feedback setting is a challenging task.
    Along these lines, prior works have proposed certain solutions for this problem; however,
    their real-world applicability has been limited. They either (1) rely on strong assumptions about the accuracy of past decisions \cite{arteaga2018selective}, 
    (2) assume that sufficiently \textit{diverse} outcomes have already been observed 
    \cite{coston2021selective} -- which is not true when past data is limited for marginalized groups, or 
    (3) classify a large number of samples positively in the beginning 
    to gather outcome information 
    \cite{bechavod2019onlineclassification} -- which results in large number of false positives in the initial iterations.
    \emph{Given these limitations, we ask if there are robust approaches for data collection and training in the partial feedback setting that achieve (a) high cumulative and iteration-wise utility and (b) low disparate impact across demographic groups?}

    \subsection{Our Contributions}
    We study the problem of data collection for accurate classification in the partial feedback setting (\cref{sec:model}).
    Our proposed framework (\cref{alg:main_algorithm}) operates in an \textit{iterative setting}, where in each iteration a set of unlabeled samples are given as input, and our framework uses the exploitation-exploration paradigm to predict the outcomes for these samples. Using the classifiers trained in all previous iterations, we first identify the ``exploit'' region, i.e., the part of the domain where accurate outcome information is available, and use the trained classifier to make predictions for samples from this region.
    The region beyond the ``exploit'' region is called the ``explore'' region and we provide \textit{a family of exploration strategies} to sample elements from this region that are also predicted as positive.
    Using the classifiers from previous iterations, we ensure a high utility over the ``exploit'' region, and by positively classifying certain samples from the ``explore'' regions, we collect outcome information about individuals and groups which would have otherwise been ignored.
    Fairness mechanisms can be incorporated in both the ``exploit'' and ``explore'' parts of our algorithm to ensure performance parity for all groups defined by given protected attributes (\cref{sec:algorithm}).
    An important aspect of our framework is the necessity to have high utility in every iteration.
    This is motivated by applications, e.g. lending settings, where high costs of erroneous decisions implicitly constrain the decision-maker to have a small number of false positives \cite{fredDelinquencyData2022,reserve2022delinquency}.
    To that end, our framework includes false discovery rate constraints guaranteeing that the expected number of false positives among the samples classified positively is small.
        Theoretically, we show that our 
        approach always satisfies the specified bound on the false discovery rate
        (\cref{thm:feasibility}).
        Furthermore, for all groups, the prediction utility is at least as high as the previous iteration (\cref{thm:fairness}); i.e., performance improvement through data collection. 
        Finally, we show that, due to exploration, the predictions of our approach converge to the predictions of an ``optimal classifier''
        (\cref{thm:convergence_finite}).
    Empirical analysis on the Adult Income and German Credit datasets further demonstrates that our proposed framework results in improved performance for all relevant groups as more data is collected (Section~\ref{sec:empirical}).
    The loss in the cumulative utility and utility per iteration due to additional exploration is minimal and the performance \mbox{disparities across protected attribute groups are reduced.}

    \subsection{Related Works}
    Certain recent works tackle the problem of partial feedback by proposing solutions that either collect additional data or modify the learning process to effectively utilize available outcome-labeled data.
    Most of these works, however, can be impractical for relevant real-world applications.
    For outcome exploration, 
    \citet{bechavod2019onlineclassification} propose
    a strategy that uses initial iterations for exploration by positively classifying all incoming samples and, in the following iterations, exploits the collected data to learn a classifier.
    Only exploring during the initial iterations, however, leads to a large number of false positives and low utility in these iterations.
    For instance, in the loan setting, a bank would be unlikely to adopt a strategy that gathers information at the cost of huge losses in certain years.
    In contrast, our framework performs both exploration and exploitation at every iteration, limiting the number of false positives per iteration.
    \citet{wei2021decisionmaking} formulates a dynamic programming framework to find a threshold-based classifier that balances exploration and exploitation. 
    \citet{kilbertus2020fair} similarly propose stochastic decision-making policies that assign a non-zero likelihood of selection 
    to every point in the domain.
    Both these works adjust the learned classification policy to implicitly explore additional samples.
    Our framework instead employs explicit data collection strategies that are more effective in improving the rate of learning (as we demonstrate in our empirical analysis in \cref{sec:empirical}).
    \citet{yangadaptive} forward a bandit-type approach that uses bounded exploration to gather additional outcome data every iteration.
    However, their method requires non-trivial parametric assumptions on the feature distribution.
    \citet{rateike2022don} develop an online process that first learns an unbiased representation of the data and then trains an online classifier over the learned representation space.
    Similar to the papers mentioned above, this approach also 
    does not employ any constraint on false positives which can lead to low utility in certain iterations when sufficient information for learning is unavailable (see \cref{sec:empirical} for empirical comparison against these methods). 
    Data collection frameworks proposed in the above works aim to \textit{eventually} collect a sufficient amount of data through exploration so that long-term prediction utility is high. However, as we discuss in the following sections, this approach often comes at the cost of low short-term or iteration-wise utility and, hence, can be inappropriate for real-world applications. 
    
    Discussions on the comparison of our work to other relevant approaches from the fields of active learning, fair classification, and classification using selective labels are presented in \cref{sec:other_related_work}.

\section{{Model, Stakeholders, and Classification}}\label{sec:model}
    
        Let $D \coloneqq \cX \times \zo \times [p]$ be a finite-sized domain, where $\cX$ is the set of features, $\zo$ is the set of labels, and $[p]$ is the set of protected attributes.
        Extensions to continuous domains and multi-class settings are discussed in \cref{sec:theoretical_results,sec:discussion} respectively.
        Let $\mu$ be the true data distribution over $D$ and $\cF \subseteq \zo^{\cX\times [p]}$ be a hypothesis class of binary classifiers.
        For any distribution $\eta$, we use $\Pr_{\eta}[\cdot]$ and $\Ex_{\eta}[\cdot]$ to denote $\Pr_{(X,Y,Z)\sim \eta}[\cdot]$ and $\Ex_{(X,Y,Z)\sim \eta}[\cdot]$ respectively.
        The goal in any classification setting
        is to find a classification policy that results in the highest \textit{utility}, where the utility definition is context-specific, and could denote measures like
        predictive accuracy or \textit{revenue}.
        To handle a wide variety of utility definitions,  
        we consider the family of utility metrics.
        
        \begin{definition}[\textbf{Utility Metrics}]\label{def:utility}
            Given tuple $\gamma \coloneqq \inparen{\gamma_{00}, \gamma_{01}, \gamma_{10}, \gamma_{11}}$, the utility  of $f\in \cF$ with respect to $\mu$ is  $\util{}_\mu(f,\gamma) \coloneqq \sum_{i,j\in \zo}
                \gamma_{ij}\cdot\Pr\nolimits_{\mu}\insquare{f(X,Z) = i,  Y = j}.$
        \end{definition}
        For a given $\gamma$, the goal of classification is to 
        solve  $\max_{f\in \cF} \util_{\mu}(f,\gamma)$.
        By using different coefficients for different kinds of predictions 
        we can capture a wide variety of utility metrics \cite{elkan2001foundations}.
        For $\gamma_{\rm acc}{=}(1,0,0,1)$, $\util{}_\mu(\cdot,\gamma_{\rm acc})$ is proportional to standard predictive accuracy and
        for $\gamma_{\rm pos} = (0,1,0,1)$, 
        $\util{}_\mu(\cdot,\gamma_{\rm pos})$ is the fraction of positive predictions.
        A metric relevant to our setting is \textit{revenue}. 
        It is the weighted sum of false positive and true positive predictions:
        given $c_1, c_2 > 0$, let $\gamma_{\rm rev} \coloneqq (0, -c_1, 0, c_2)$, then ${\rm revenue}_{c_1, c_2, \mu}(f) \coloneqq \util{}_\mu(f, \gamma_{\rm rev}) \cdot (\textrm{number of samples})$.
        Here, $c_1$ represents the absolute value of loss incurred for making a false positive error and $c_2$ represents profit acquired for a true positive prediction. %
        
        Group-specific performance and classifier \textit{fairness} can be measured with conditional utility: for group $z$, define 
        $\util{}_\mu(f,\gamma, z) \coloneqq \util{}_{\mu \mid Z=z}(f,\gamma)$.
        As an example, for $\gamma_{\rm tpr} = {\textstyle(0, 0, 0, 1/\Pr[Y{=}1 \mid Z{=}z])}$, $\util{}_\mu(f,\gamma_{\rm tpr}, z)$ denotes the true positive rate (TPR) for group $z$.
        Performance disparity of $f$ across groups $z_0, z_1 {\in} [p]$ can then be quantified as $|\util{}_\mu(f,\gamma, z_0) - \util{}_\mu(f,\gamma, z_1)|$, i.e., absolute difference between group-wise utilities.
        With $\gamma = \gamma_{\rm pos}$, this denotes acceptance rate disparity (or \textit{statistical rate} \cite{celis2019classification}) and with $\gamma = \gamma_{\rm tpr}$, this denotes TPR disparity.

    \subsection{Partial Feedback, False Discovery Rate, and Optimal Offline Classifier}
        We consider the \textit{partial feedback} setting, where true outcome labels are only observed for samples that were positively classified in the past.
        While the usual goal of classification is to ensure high predictive accuracy, in the partial feedback setting, there is an additional goal to gather information about unobserved samples.
        A trivial approach to data collection is to positively classify all samples and then observe the true outcomes.
        While this would lead to rich data collection, it will also have poor classifier utility.
        Applications involving high-stakes decisions, e.g. credit lending, usually attempt to make as few false positive predictions as possible.
        This is because the losses due to false positives are often much larger than profits from true positives
        (in the $\textrm{revenue}_{c_1, c_2}(\cdot)$ metric defined above, this is characterized by $c_1 > c_2$).
        For such applications, it is necessary to limit the number of false positive errors made which can be encoded using the \textit{false discovery rate}. %
        \begin{definition}[\textbf{False-discovery Rate Constraint}]\label{def:fdrConstraint}
            For any $\alpha\in (0,1]$, $f\in \cF$ is said to satisfy $\alpha$-false-discovery rate constraint (or $\alpha$-FDR) if
            $\Pr_{\mu}\insquare{Y=0\mid f(X,Z)=1} \leq \alpha.$
        \end{definition}
        FDR captures the fraction of false positives among the samples classified positively.
        When losses associated with false positives are larger in magnitude than the profits associated with true positives, having a high FDR can lead to potentially negative utility.
        Hence, using an appropriate \textit{non-trivial} FDR constraint in our framework can ensure that utility per iteration is lower bound by a positive amount. %
        For a given $\alpha$ and $\gamma$, the goal of classification with partial feedback is to converge to the optimal offline classifier $\optOffline^\alpha$, where
        $\optOffline^\alpha$ is the classifier with
        the maximum utility with respect to true distribution $\mu$ subject to $\alpha$-FDR constraint:
        \begin{align*}
            \optOffline^\alpha %
            ~\coloneqq~~ 
            &\argmax\nolimits_{f\in \cF}\ \util(f,\gamma),
            \quad\text{such that,}\quad
            \text{$f$ satisfies $\alpha$-FDR}.%
            \yesnum\label{eq:optOffline}%
        \end{align*}

        \subsection{Stakeholders and Iterative Model}
                Our setting is iterative: At each iteration $t \in \{1,2,\dots\}$, an institution 
                needs to make predictions about a (new) set of $n\in \N$ individuals.
                E.g., suppose each year a bank must make predictions for a fresh set of loan applicants.
                Before the first iteration, the institution had a decision-making process in place that it used to make past decisions.
                This could either be just human decision-makers 
                or a
                classifier $f_0$. %
                We assume that the labeled samples $L_0$ (i.e., samples predicted positively in the past) and unlabeled samples $U_0$ from the past (i.e., samples predicted negatively in the past) are available.
                If the past predictions were made by humans, then one can train a classifier $f_0$ to simulate human decision-making (by simulating the partition between $L_0$ and $U_0$).
                Importantly, we make the following minimal assumptions on $f_0$.
                \begin{assumption}\label{assump:f0_is_feasible}
                    \textit{We assume $f_0\in \cF$ is $(\alpha,\lambda)$-feasible: a classifier $f$ is 
                    $(\alpha,\lambda)$-feasible if 
                        (1) (feasibility) $f$ satisfies the $\alpha$-FDR constraint; and 
                        (2) (positive selection rate) there exists a constant $\lambda\in [0,1]$ such that $\Pr_{\mu}[f(X,Z){=}1]\geq \lambda$. }
                \end{assumption} 
                We require \cref{assump:f0_is_feasible}~(1) to ensure that the $\alpha$-FDR constraint is satisfied in the first few iterations (when our predictions are similar to $f_0$) and \cref{assump:f0_is_feasible}~(2) to prove concentration bounds on the FDR of $f_0$.
                We expect \cref{assump:f0_is_feasible} to be satisfied in real-world applications.
                For instance, consider credit lending. 
                The first requirement in \ref{assump:f0_is_feasible} holds as bank policies usually require them to limit the fraction of applications that are erroneously accepted \cite{fredDelinquencyData2022}, which implies a constraint on the false discovery rate of their decision-making policy.
                The second requirement simply ensures policies making near-zero positive classifications (that are not practically relevant) are ruled out.\footnote{In the absence of this requirement, a trivial approach to reducing the number of false positives is to make zero positive classifications, which is undesirable in many applications.}

                The features-label pairs 
                at the $t$-th iteration correspond to a set $S_t$ of $n$ i.i.d.\ samples from $\mu$.
                However, the institution only has access to features $X_t \coloneqq S_t|_\cX$ and not the labels.
                After making predictions $\hy_x$ for each $x\in X_t$, the institution observes the labels of all positively classified samples, i.e., the institution observes $\inbrace{y\mid (x,y)\in S_t \text{ and } \hy_x=1}$.
                This process ``partitions'' $S_t$ into a labeled set $L_t \coloneqq \inbrace{(x,y) \mid (x,y) \in  S_t \text{ and } \hy_x=1}$ and an unlabeled set $U_t \coloneqq \inbrace{x\mid (x,y) \in S_t \text{ and }\hy_x = 0}$.
                Sets $\inparen{L_i,U_i}_{i=1}^t$ can be used for prediction in future iterations.
                Note that the predictions $\hy$ should satisfy the $\alpha$-FDR constraint so that the utility per iteration is high.
                    Prior works do not satisfy this constraint and, hence, often have low iteration-wise utility (as observed in Section~\ref{sec:empirical}).

\begin{figure}
    \centering
    \includegraphics[width=\linewidth]{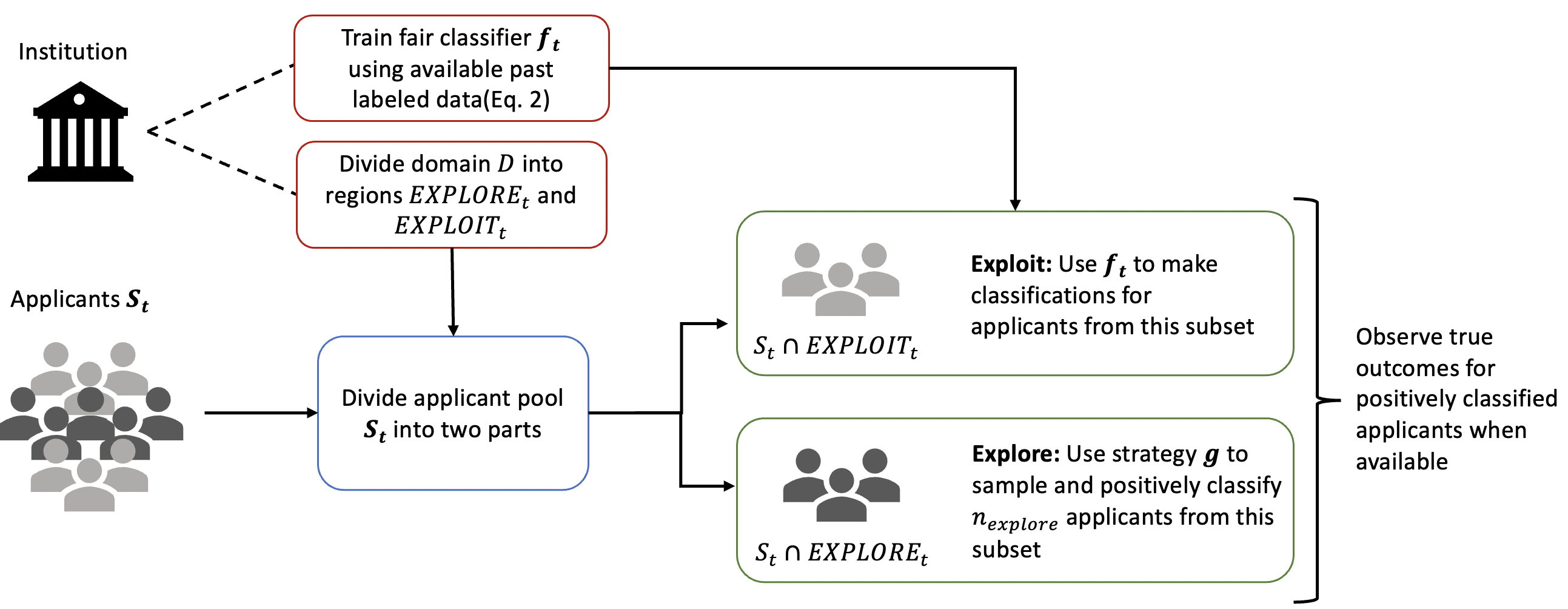}
    \caption{Pipeline of the process undertaken at time-step $t$ to classify unlabeled samples $S_t$. The institution learns a classifier $f_t$ using past labeled data. It also creates exploitation-exploration partitions to decide which elements in $S_t$ will be classified using $f_t$ and which elements will compose the exploration set, over which the exploration strategy $g$ will be employed.}
    \label{fig:process}
\end{figure}

\section{Our Framework}\label{sec:algorithm}
    As mentioned earlier, our approach for simultaneous data collection and prediction is designed to handle the partial feedback setting.
    To do so, at each iteration $t$,
    we partition the domain $D$ into two regions: $\expl{}_t$ (initialized to be empty before the first iteration) and $\expr{}_t$ (defined as $D \setminus \expl{}_t$).
    Region $\expl{}_t$ contains all points for which {outcome labels have been observed \textit{sufficiently} many times in the past}: concretely, each point $(x,z)$ has a weight $w_t(x,z)$--which is proportional to the number of times its outcome has been observed in the previous iterations--and $(x,z)$ is included in $\expl{}_t$ if $w_t(x,z)>\tau$ (where $\tau$ is a fixed pre-specified threshold).
    For any $(x,z)$, $w_t(x,z)$ never decreases from one iteration to the next, and hence, points are only added to the exploitation region and removed from the exploration region.
    Figure~\ref{fig:process} demonstrates our workflow.
    The pseudocode of our approach is presented in \cref{alg:main_algorithm} and we next describe its various components.\footnote{Note that occurrences of $D$ in \cref{alg:main_algorithm} can be replaced by $S_t$ if the domain $D$ is extremely large in practice.}

 \begin{algorithm}[t!] 
            \caption{~~~~Data Collection and Prediction Framework}\label{alg:main_algorithm}
            
            \textbf{Input:} 
            Hypothesis class $\cF$ with $f_0\in \cF$, $\alpha>0$, 
            labeled dataset $L_0$, and 
            unlabeled data stream $S_0,S_1,\dots, S_T$.
            Constants $\alpha_{\rm exploit} \in (0, \alpha)$,
            $\eps, \tau, \lambda\in (0,1]$, and 
            exploration strategy $g$.

        \begin{algorithmic}[1]
        
            \STATE Initialize $\expl_1=\emptyset$ and $\expr_1=D$ 
            \FOR{$t=1,2,\dots, T$}
                \vspace{1mm}
                \item[]  \textit{Learn:}
                \begin{ALC@g}
                    \STATE \text{If } $t=1$, set $f_t = f_0$ (since $\expl_1=\emptyset$), otherwise

                    \vspace{-4mm}
                    \begin{align*}
                        \textstyle &f_t\quad = \quad
                        \arg\max\nolimits_{h\in \cF} \ \ \textstyle \util_{\eta_w}\inparen{h,\gamma},\yesnum\label{prog:main_algorithm}\\
                        \textstyle
                        &\qquad\qquad\st,~~~~ \ 
                            \textstyle \Pr\nolimits_{\eta_w}\insquare{h=1}
                            \geq \lambda - \eps
                            \quad\text{and}\quad 
                            \Pr\nolimits_{\eta_w}\insquare{h \neq  y\mid h=1}\leq \alpha_{\rm exploit} + \eps.
                    \end{align*}
                    
                    Where $\eta_w(x,y,z)$ is a density proportional to the product of three terms: (1) the indicator $\mathds{I}\insquare{(x,z)\in \expl_t}$, (2) the probability that $(x,z)$ is in $\bigcup_{i=0}^{t} S_i$ and (3) the probability that $(x,z)$ is in $\bigcup_{i=0}^{t-1} L_i$. %
                \end{ALC@g}
                
                \vspace{2mm}
                \item[] \textit{Exploit:}
                \begin{ALC@g} 
                    \STATE $\wh{L}_t := \inbrace{(x, z) \in S_t\cap \textsc{Exploit}_t \mid f_t(x,z)=1}$ 
                    \STATE $n_{\rm exploit} = \sabs{\wh{L}_t}$ 
                \end{ALC@g} 
                
                \vspace{2mm}
                \item[] \textit{Explore:}
                \begin{ALC@g}
                    \STATE Set $n_{\rm explore}=(\alpha-\alpha_{\rm exploit}-\eps) \cdot n_{\rm exploit}/(1-\alpha)$ 
                    \STATE For all $(x,z) \in S_t\cap \expr{}_t$, set $p(x,z) \propto g(x, z; f_t)$
                    \STATE Sample $n_{\rm explore}$ points from $S_t\cap \expr{}_t$ using distribution $p$ and add them to $\wh{L}_t$
                \end{ALC@g} 
                
                \vspace{2mm}
                \item[] \textit{Observation and Region Update:}
                \begin{ALC@g}
                    \STATE For each $(x,z) \in \wh{L}_t$, observe label $y$
                    \STATE Create $L_t = \sinbrace{(x,z,y) {\mid} (x,z) \in  \wh{L}_t, y \text{ is $(x,z)$'s observed outcome}}$
                    \STATE Initialize $\expl_{t+1} = \expl_{t}$
                    \STATE For each $(x,z)\in \expr_{t-1}$, such that $w_{t+1}(x,z)=\sum_{1\leq i\leq t} g(x,z; f_i)>\tau$, add $(x,z)$ to $\expl_{t+1}$
                    \STATE Set $\expr_{t+1}=D\setminus \expl_{t+1}$
                \end{ALC@g}
            \ENDFOR{}
        \end{algorithmic}
    \end{algorithm}

    Given FDR parameter $\alpha$ as input, fix any $\alpha_{\rm exploit}\in (0,\alpha)$. 
        At iteration $t$, \cref{alg:main_algorithm} receives an unlabeled dataset $S_t$.
        
        \begin{description}
            \item[\textbf{Learning (Step 3).}] 
            First, \cref{alg:main_algorithm} learns a classifier $f_t$ that maximizes the utility over $\expl_t$ subject to satisfying $\alpha_{\rm exploit}$-FDR constraint and making sufficiently many positive predictions.
            One difficulty in learning $f_t$ is that the empirical distribution over $\expl{}_t$ may not be an unbiased estimate of the true distribution $\mu$ (as certain samples are over-represented due to past-decisions).
            We correct this by optimizing the utility with respect to a re-weighted distribution $\eta_w$ which ensures unbiasedness with respect to $\mu$.
            Intuitively, $\eta_w$ is a product of three terms: an indicator ensuring the support of $\eta_w$ is $\expl_t$, 
            (b) 
            $\mu(x,z)$, and 
            (c) 
            $\Pr_\mu[Y{=}y{\mid}X{=}x,Z{=}z]$.
            If these terms are known exactly, then one can show that, by chain rule of probability, $\eta_w$ is the same as $\mu$ on $\expl_t$.
            \cref{alg:main_algorithm}  uses estimates of these terms, which we show is good enough in our theoretical analysis (e.g., Eq.~\ref{eq:generalizationBound:main} generalization bound).
            This optimization can be solved using standard cost-sensitive classification methods (\cref{sec:implementation_details}).
            \item[\textbf{Exploitation (Step 4-5).}]    
            Next, \cref{alg:main_algorithm} uses $f_t$ to predict the labels for samples in $S_t\cap \expl_t$.
            Further, by design, if $n_{\rm exploit}$ samples are positively predicted in this step, then there are at most $n_{\rm exploit}\cdot \alpha_{\rm exploit}$ false positives.
            \item[\textbf{Exploration (Steps 6-8).}] 
            The exploration region consists of samples for which sufficient outcome information is not available.
            To determine which samples from $\expr_t$ region are positively predicted, we use function $g$, which we call the \textit{exploration strategy}.
            Concretely, \cref{alg:main_algorithm} draws a ``certain'' number of elements from $\expr{}_t$, with sampling probability of any point $(x,z)$ being proportional to $g(x,z; f_t)$ and predicts a positive label for these elements.
            \item[\textbf{Observation and Region Update (Steps 9-13).}] Finally,
                \cref{alg:main_algorithm} observes the true outcome labels of positively-predicted samples. %
                In practice, this observation step entails checking if each positively classified individual satisfies certain requirements, e.g., whether a loan is paid back within two years (see \cref{sec:discussion} for more discussion on this point).
                Before the next iteration, we also update the exploitation region to include points $(x,z)$ whose weight $w_{t+1}(x,z)=\sum_{1\leq i\leq t} g(x,z; f_i)$ now exceeds $\tau$ (observe that $w_{t+1}(x,z)$ is proportional to the number of times the label of $(x,z)$ has been observed in the first $t$ iterations).
        \end{description}
        Two aspects of the exploration step that need further description are (a) the number of samples, $n_{\rm explore}$, from $\expr{}_t$ region that are positively classified, and (b) the exploration strategy $g$.
        Regarding (a),
        \cref{alg:main_algorithm} sets $n_{\rm explore}{=}n_{\rm exploit}\cdot \nfrac{(\alpha-\alpha_{\rm exploit})}{1-\alpha}$.
        This choice allows us to control the number of false positives and satisfy $\alpha$-FDR.
        Roughly, even if all $n_{\rm explore}$ samples from the explore region are false positives, the combined number of false positives from the exploration and exploitation steps is at most $\alpha \cdot n_{\rm exploit}\cdot\nfrac{(1-\alpha_{\rm exploit})}{1-\alpha}$, which ensures $\alpha$-FDR since the total positive predictions is $n_{\rm exploit}\cdot\nfrac{(1-\alpha_{\rm exploit})}{1-\alpha}$.
        See \cref{thm:feasibility} for formal proof of this feasibility claim.
        
        Regarding (b), 
        \cref{alg:main_algorithm}'s only requirement for $g$ is that it take positive values so that all points have a positive probability of being observed.
        When no information is available about the samples in the $\expr{}_t$, the obvious choice for $g$ is the uniform distribution.
        {However, that is rarely the case in practice: 
        while $\expl{}_t$ and $\expr{}_t$ could differ in group composition and will likely belong to different underlying feature-label distributions, there can be some similarities across different groups that would 
        allow classifiers trained on $\expl_t$ to be partially predictive on $\expr_t$.}
        E.g., say in a loan application setting, 
        suppose two applicants from different groups have very high credit scores. Even if the $\expl_t$ contains data from only one group, 
        information about the high credit score applicant in one group can be used to judge that high credit score applicants from all groups have low default risk.
        In such cases, classifier $f_t$ can be utilized for exploration as well by, say, choosing $g^{\rm clf}(x,z; f_t)\propto \beta+(1-\beta)f_t(x,z)$.
        Parameter $\beta \in [0,1]$ will depend on the expected accuracy of $f_t$ over $\expr_t$. %
        
        \smallskip\noindent\textbf{Fairness in exploration.}
        While exploration strategy $g = g^{\rm clf}$ can allow for improved exploration utility, it can exacerbate social biases if $f_t$ is biased towards favoring certain groups. %
        To reduce performance disparity across groups, it is therefore important to explore the outcomes of individuals from marginalized groups at an increased rate.
        This can be accomplished by choosing $g$ in a manner that takes into account the proportions of different groups.
        E.g., $g^{\rm fair}(x,z; f_t) \propto g^{\rm clf}(x,z; f_t) \cdot \Pr_\mu\insquare{Z=z {\mid} (X, Z)\in \expr_t}$ would use classifier output to improve utility while ensuring that every group's selection rate is close their proportion in $\expr_t$.
        Hence, groups that are under-represented in $\expl_t$, 
        compared to their population proportion, will be explored at a higher rate.
        See \cref{sec:discussion} for other choices and discussion of $g$.
        
        \smallskip\noindent\textbf{Fairness in exploitation.}
            A common approach to mitigate biases in classification is to use constraints during learning that require performance disparity across groups to be small \cite{celis2019classification}.
            Our framework can incorporate such fairness mechanisms by constraining the classifier trained in Step~3 using any common fairness metric.
            Fairness constraints in exploitation may be necessary when \cref{alg:main_algorithm} is implemented using a biased data source (which is the case in \cref{sec:empirical} simulations).
            In such cases, despite re-weighting, the distribution used for training in Step~3 of \cref{alg:main_algorithm} can still misrepresent marginalized groups.
            However, just using fairness constraints during exploitation is not sufficient as it only ensures fairness over $\expl_t$ (which may not represent the entire domain).
            Hence, it should be used along with exploration fairness.
            As observed in \cref{sec:empirical}, group disparity is smallest when fairness is incorporated in both exploit steps and explore steps.

\section{Theoretical Results}\label{sec:theoretical_results}
    We next present the theoretical guarantees of our framework; all proofs are provided in \cref{sec:proofs}.
    The sample complexities in our results depend on ``how uniform $g$ is.''
    One way to capture this is via the following parameter:
    \begin{align*}
        \sigma \coloneqq \frac{{\min_{(x,z) \in D} g(x,z)}}{{\sum_{(x,z)\in D} g(x,z)}}.
        \yesnum\label{def:definitionSigma}
    \end{align*}
    $\sigma$ is non-zero as $g>0$.
    It is minimized when $g$ approaches 0 at some $(x,z)\in D$, and it is maximized when $g$ is the uniform distribution over $D$.
    Our first result shows that \cref{alg:main_algorithm} satisfies the specified FDR constraint. %
    \begin{restatable}[\textbf{Feasibility with respect to FDR constraint}]{theorem}{thmFeasibility}\label{thm:feasibility}
        Suppose $f_0$ is $(\alpha,\lambda)$-feasible (\cref{assump:f0_is_feasible}). 
        For any $\eps,\delta,\tau \in (0,1]$, \cref{alg:main_algorithm} satisfies the following at every iteration $t$:
        If $n\geq \abs{D}\cdot\poly\big(\nfrac{1}{\lambda},$ $\nfrac{1}{\tau}, \nfrac{1}{\min\inbrace{\eps,\alpha-\eps}}\big)\cdot \log\inparen{\nfrac{\abs{D}}{\sigma\delta}}$,
        then the predictions made in the $t$-th iteration satisfy the $\alpha$-FDR constraint with probability at least $1-\delta$ with respect to 
        the randomness in $S_1,S_2,\dots,S_t$ and \cref{alg:main_algorithm}.
    \end{restatable}
    \cref{thm:feasibility} holds for any exploration strategy $g$ which takes positive values. %
    In line with other works using exploration for data collection (e.g., \citet{wei2021decisionmaking}), \cref{thm:feasibility} requires $n$ to be linear in $\abs{D}$. 
    This dependence can be improved by making additional assumptions on $\mu$:
    for instance, if $\mu$ is ``smooth,'' then one can reduce the sample complexity from $\abs{D}$ to $C$, where $C$ is the minimum number clusters that achieve a ``high prediction accuracy'' (see Theorem 19.3 of \citet{shalev2014understanding} and \cref{sec:additionalDiscussion}).

    Our next result shows that the group-wise utility of the classifiers learned by our framework increases at every iteration.
    The results holds for hypothesis classes $\cF$ where each hypothesis $f\in \cF$ is a tuple of $p$ ``classifiers'' $f=(f_1,f_2,\dots,f_p)$, one for each group.
    Here, each classifier $f_z$ belongs to some \textit{base hypothesis class} $\cB\subseteq \zo^{\cX}$ (e.g., set of linear classifiers) and we say $\cF$ is \textit{derived} from $\cB$. 

    \begin{restatable}[\textbf{Fairness: Improvement in group-wise utility}]{theorem}{thmFairness}\label{thm:fairness}
        Suppose $f_0$ is $(\alpha,\lambda)$-feasible (\cref{assump:f0_is_feasible}) and $\cF$ is derived from $\cB\subseteq \zo^{\cX}$.
        For any $\eps,\delta,\tau\in (0,1]$ and tuple $\gamma$, \cref{alg:main_algorithm} 
        satisfies the following at every iteration $t$ and $z\in [p]$:
        If $n\geq \abs{D}\cdot\poly\big(\nfrac{1}{\lambda},\nfrac{1}{\tau}, \nfrac{1}{\min\inbrace{\eps,\alpha-\eps}}\big)\cdot \log\inparen{\nfrac{\abs{D}}{\sigma\delta}}$,
        then with probability at least $1-\delta$,
        \[
            \util_{\mu,t}(f_t,\gamma, z)
            \geq 
             \max_{0\leq i\leq t-1} \util_{\mu,t}(f_i,\gamma,z)-\eps.
        \]
        Where $\util_{\mu,t}(f,z)$ is the utility of $f$ over draws $(X,Y,Z)\sim \mu$ conditioned on $(X,Z) \in \textsc{Exploit}_t$ and $Z = z$.
        The randomness at the $t$-th iteration is with respect to the randomness in $S_1, \dots, S_t$ and \cref{alg:main_algorithm}.
    \end{restatable}
    \cref{thm:fairness} holds for any $\util$ and shows that, with high probability, $f_t$ achieves a higher utility for each group than any previously learned classifier.
    At the $t$-th iteration, the utility in \cref{thm:fairness} is measured with respect to $\textsc{Exploit}_t$ (since $f_t$ is only used to make predictions for samples in $\textsc{Exploit}_t$).%
    
    Our final theoretical result shows that $f_t$'s utility converges to $\optOffline$'s utilitys as $t{\to}\infty$ 
    since, for any $t$, $f_t$ is ``accurate'' on  $\textsc{Exploit}_t$ and, 
    as $t{\to}\infty$, $\textsc{Exploit}_t$ converges to $D$.
    Intuitively, this is true because \cref{alg:main_algorithm} observes the true labels of all samples with a positive probability. 
    Convergence-rate for group $z$ depends on $\tau$, $\alpha_{\rm explore}$, and $\sigma(z)$.
    Where $\sigma(z) \coloneqq \sfrac{{\min_{x \in \cX} g(x,z)}}{\sinparen{\sum_{(x,z)\in D} g(x,z)}}$ is the fraction of mass that function $g$ assigns to group $z$.

    \begin{restatable}[\textbf{Group-wise convergence to $\optOffline^\alpha$}]{theorem}{thmConvergenceFinite}\label{thm:convergence_finite}
        Suppose $f_0$ is $(\alpha{-}\eps,\lambda)$-feasible.
        For any $\alpha,\eps,\delta,\tau\in (0,1]$ and $\alpha_{\rm exploit} = \alpha - \eps$, $\alpha_{\rm explore} = \eps,$ \cref{alg:main_algorithm} satisfies the following: 
        if
            $t\geq \nfrac{1}{\sigma(z)}$ and 
            $n\geq \abs{D}\cdot\poly\big(\nfrac{1}{\lambda},\nfrac{1}{\tau}, \nfrac{1}{\min\inbrace{\eps,\alpha-\eps}}\big)\cdot \log\inparen{\nfrac{\abs{D}}{\sigma\delta}}$
        then with probability at least $1-\delta$, the utility of the classifier $f_t$ learned by the framework in the $t$-th iteration is at least as large as the utility of $\optOffline^{\alpha}$ on the samples in the $z$-th group drawn from $\mu$, i.e., %
        \[
            \util_\mu\inparen{f_t,\gamma,z}
            \geq 
            \util_\mu\inparen{
                \optOffline^{\alpha}, \gamma, z
            } - \eps.
        \]
        Where the randomness at the $t$-th iteration is with respect to the randomness in $S_1, S_2,\dots, S_t$ and \cref{alg:main_algorithm}.
    \end{restatable}
    The convergence rate for the $z$-th group increases with $\sigma(z)$ and, hence, choosing $g$ that explores samples in $z$-th group with higher probability improves the convergence rate on the $z$-th group.
    This may be desirable in some contexts to address historical biases (see \cref{sec:algorithm} and \cref{sec:discussion}).
    Finally, the following convergence bounds are a corollary of \cref{thm:convergence_finite}:
    if
            $n\geq \abs{D}\cdot$ $\poly\inparen{\nfrac{1}{\lambda},\nfrac{1}{\tau}, \nfrac{1}{\min\inbrace{\eps,\alpha-\eps}}}\cdot \log\inparen{\nfrac{\abs{D}}{\sigma\delta}}$
            and 
            $t\geq \nfrac{1}{\sigma},$
    then with probability at least $1-\delta$, $ 
        \util_\mu\inparen{f_t,\gamma}$ $
            \geq 
            \util_\mu\inparen{
                \optOffline^{\alpha}, \gamma} - \eps.
    $
    I.e., $f_t$'s utility is at least as large as that of $\optOffline^{\alpha}$ on the samples drawn from $\mu$. %

    \smallskip 
    \noindent\textbf{Key proof technique.}
        The technical core of our 
        analysis is a generalization bound (\cref{lem:conc_inequality}) showing that the reweighted distribution $\eta_w$ in Step~3 of \cref{alg:main_algorithm} is a good approximation of $\mu$ on $\expl_t$.
        We show that
        if $n\geq \abs{D}\cdot\poly\big(\nfrac{1}{\lambda},\nfrac{1}{\tau}, \nfrac{1}{\min\inbrace{\eps,\alpha-\eps}}\big)\cdot \log\inparen{\nfrac{\abs{D}}{\sigma\delta}}$, then 
        at any iteration $t$ and for any bounded function $h\colon \zo\times \zo\times [p]\to[0,1]$ the following holds:
        with probability at least $1-\delta$, for all $f\in \cF$
        \[
                    \bigg|
                        \Ex_{\eta_w\mid (X,Z)\in \expl_t}\insquare{h(f(X,Z),Y,Z)}
                        ~~-~~
                        \Ex_{\mu\mid (X,Z)\in \expl_t}\insquare{h(f(X,Z),Y,Z) }
                    \bigg|
                    ~~\leq~~ O(\eps).
                \yesnum\label{eq:generalizationBound:main}
        \]
        At any $t$, $\eta_w$ is a product of three terms (see {Step~3} of \cref{alg:main_algorithm}): (a)
        an indicator ensuring the support of $\eta_w$ is $\expl_t$, 
        (b) an estimate of $\mu(x,z)$, and 
        (c) an estimate of $\Pr_\mu[Y{=}y{\mid}(X,Z){=}(x,z)]$.
        If the estimates in (b) and (c) are exact, then Eq. \ref{eq:generalizationBound:main} follows.
        We prove that, with probability $ \geq 1{-}\delta$, both estimates are nearly-correct for all $(x,z)  \in  \expl{}_t$ with $\mu(x,z) > \nfrac{\eps}{\abs{D}}$.
        This implies that with probability $ \geq 1-\delta$, 
        $\eta_w(x,y,z)\in (1\pm O(\eps)) \mu(x,y,z)$ for any $(x,z)\in \expl_t\setminus R$ and $y\in \zo$, where $R \coloneqq \inbrace{(x,z)\mid \mu(x,z)\leq \nfrac{\eps}{\abs{D}}}$.
        That is, $\eta_w$ is a good approximation of $\mu$ on $\expl_t\setminus R$ -- Eq. \ref{eq:generalizationBound:main} follows as $\mu(R)\leq \eps$.
        The proof of estimate (b)'s accuracy is via the Chernoff-bound.
        For (c), if we show that
        $\bigcup_{i=0}^{t-1} L_i$ has at least $k_0\coloneqq \poly\inparen{\nfrac{1}{\eps}}\cdot\log\inparen{\nfrac{\abs{D}}{\delta}}$ copies of each $(x,z)\in \expl_t\setminus R$, then bound on estimate (c) follows.
        Fix any $(x,z)\in \expl_t\setminus R$.
        Let $N_i$ be the number of copies of $(x,z)$ in $L_i$.
        If $N_1, N_2, \dots$ are independent, then existing concentration inequalities imply the claim.
        The main challenge is that independence may not hold as \cref{alg:main_algorithm}'s choices depend on past observations.
        To overcome this, we show that $N_1,\dots, N_T$ are mutually independent, where $T$ is the last iteration when $(x,z)$ is in the exploration region (as before, the observed labels for $(x,z)$ do not affect $\eta_w$).
        \cref{sec:proofs} has a detailed overview and proof. %

\begin{table}[t]
    \centering
    \small
    \caption{Comparison of all methods on the Adult (race) and German (gender) datasets.
    We report the avg. revenue per iteration (standard error in brackets), avg. FDR, and avg. acceptance rate disparity (statistical rate).
    Parameter details are provided in Figure~\ref{fig:results_explore_exploit_race_only_ours}, \ref{fig:results_explore_exploit_only_ours_german} captions.
    }
    \begin{tabular}{lcccc}
    \toprule
    & \multicolumn{4}{c}{Adult - Protected attribute: Race} \\    
    Method & \specialcell{Revenue\\(in thousands)} & FDR  & Statistical Rate  & \specialcell{TPR Disparity} \\
    \midrule
    Algorithm~\ref{alg:main_algorithm}- no fairness constraint & 71.5 (12.6)  & .15 (.02)  & .02 (.02)  & .08 (.05) \\ 
    Algorithm~\ref{alg:main_algorithm}- only exploit fairness & 73.0 (12.0)  & .15 (.02)  & .02 (.01)  & .08 (.05) \\ 
    Algorithm~\ref{alg:main_algorithm}- only explore fairness & 74.7 (12.0)  & .15 (.02)  & .03 (.02)  & .07 (.06) \\ 
    Algorithm~\ref{alg:main_algorithm}- both fairness constraints & 74.1 (12.2)  & .15 (.02)  & .01 (.01)  & .06 (.04)\\ 
    \midrule
    Baseline - \textsc{Opt-offline}  & 75.7 (9.9)  & .14 (.02)  & .03 (.02)  & .12 (.08) \\ 
    Baseline - \textsc{Fair-clf}  & 71.1 (1.4)  & .14 (.02)  & .03 (.02)  & .11 (.08)  \\ 
    \textsc{Kilbertus et al.}  & 44.9 (14.1)  & .12 (.03)  & .03 (.02)  & .09 (.07) \\  
    \textsc{Yang et al.}   & -14.7 (16.4)  & .35 (.07)  & .12 (.04)  & .12 (.06) \\        
    \textsc{Rateike et al.}   & -17.2 (8.1)  & .12 (.01)  & .02 (.01)  & .02 (.01) \\        
    \bottomrule
    \end{tabular} 

    \vspace{3mm} 

    \begin{tabular}{lcccc}
        \toprule
        & \multicolumn{4}{c}{German - Protected attribute: Gender} \\    
        Method & \specialcell{Revenue\\(in thousands)} & FDR  & Statistical Rate  & \specialcell{TPR Disparity}  \\
        \midrule
    Algorithm~\ref{alg:main_algorithm}- no fairness constraint & 6.4 (4.2)  & .11 (.07)  & .07 (.06)  & .08 (.07) \\ 
    Algorithm~\ref{alg:main_algorithm}- only exploit fairness &  8.3 (4.1)  & .11 (.05)  & .04 (.04)  & .06 (.05) \\ 
    Algorithm~\ref{alg:main_algorithm}- only explore fairness & 9.4 (4.3)  & .11 (.05)  & .07 (.05)  & .08 (.07) \\ 
    Algorithm~\ref{alg:main_algorithm}- both fairness constraints &  8.8 (4.1)  & .11 (.05)  & .05 (.04)  & .06 (.06)  \\ 
    \midrule
    Baseline - \textsc{Opt-offline}  &  9.7 (2.7)  & .15 (.04)  & .15 (.04)  & .15 (.05)  \\ 
    Baseline - \textsc{Fair-clf}  &  8.7 (3.9)  & .13 (.05)  & .05 (.05)  & .06 (.06) \\ 
    \textsc{Kilbertus et al.}  &  9.7 (5.4)  & .24 (.03)  & .11 (.09)  & .09 (.08) \\  
    \textsc{Yang et al.}   & -1.7 (9.7)  & .29 (.04)  & .08 (.14)  & .06 (.13) \\        
    \textsc{Rateike et al.}   & 2.2 (0.8) & .25 (.01)  & .04 (.02)  & .05 (.02) \\        
    \bottomrule
    \end{tabular} 
    \label{tab:overall_adult_race}
\end{table}

\section{Empirical Results}\label{sec:empirical}

\subsection{{Adult Income Dataset}} 
\label{sec:adult}
We first evaluate our framework over the new Adult Income dataset
which contains demographic and financial data of around 251k individuals from California \cite{ding2021retiring}. 
The task is to predict whether the annual income of an individual is above \$50k
and can be employed in lending settings to determine credit risk.
We use race and gender as protected attributes. The dataset is almost evenly divided with respect to gender, and for race, we limit the dataset to White (93\%) and Black/African-American (7\%) individuals.
Feature and pre-processing details are provided in Appendix~\ref{sec:empirical_other}. We present results for race in this section and results for gender are deferred to Appendix~\ref{sec:empirical_other}.
We test our algorithm over 40 iterations and the dataset is randomly split into 40 equal parts.
The first part, denoted by $S_0$, is used to construct the initial dataset, and the $i$-th part is the input for the $i$-th iteration. 
We create an initial dataset using $S_0$ that simulates real-world biased data settings.
To do so, $S_0$ is divided into a labeled subset $L_0$ and an unlabeled subset $U_0$,
with $L_0$ containing 90\% samples with class label 1 and only 10\% samples with class label 0, and $U_0 = S_0{\setminus} L_0$.
$L_0$ represents decisions made by prior biased mechanisms (e.g., biased human decision-makers) that primarily accepted samples for which accurate decisions could be made.
An initial logistic classifier ($f_0$) is trained to simulate the prior decisions.
Assigning samples in $L_0$ with a dummy label of 1 and samples in $U_0$ with a dummy label of 0, we train $f_0$ to simulate past decision boundaries. 

We use constrained logistic regression with adjusted thresholds to compute $f_t$ at any iteration $t$.
Utility is measured using the $\textrm{revenue}_{c_1,c_2}(\cdot)$ metric with $c_1 = 500$ and $c_2 = 200$.
The FDR constraint parameter $\alpha = 0.15$.
Performance for different $\alpha$ and implementation details for the optimization program are provided in \cref{sec:empirical_other}.
We perform 50 repetitions using a random dataset split in each repetition.\footnote{Link to code: {\small\url{https://github.com/vijaykeswani/Fair-Classification-with-Partial-Feedback/}}}

\begin{figure*}[t]
    \centering
    \includegraphics[width=\linewidth]{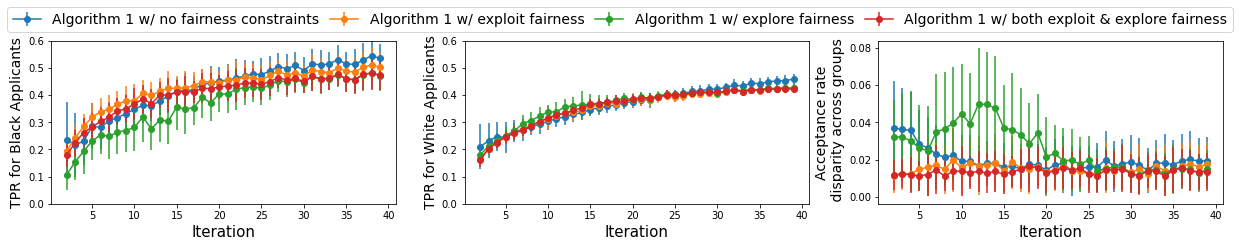} 
    \caption{\small Iteration-wise performance of Algorithm~\ref{alg:main_algorithm} (with explore and/or exploit fairness) on Adult (protected attribute is race). 
    Parameters $\alpha = 0.15$ with $\alpha_{\textrm{exploit}}  =  0.075 {\cdot} t^{0.2}$ and $\alpha_{\textrm{explore}}  =  \alpha  -  \alpha_{\textrm{exploit}}$, $\tau = 0.5, \lambda  = 0, \epsilon = 10^{-3}$.
    }
    
    \label{fig:results_explore_exploit_race_only_ours} 
\end{figure*}

\noindent
\textbf{Fairness constraints.} 
As fairness can be incorporated in both the exploration and exploitation components of Algorithm~\ref{alg:main_algorithm}, we obtain four variants of our algorithm: \textit{(a) no fairness constraints, (b) only exploit fairness, (c) only explore fairness}, and \textit{(d) both exploit and explore fairness constraints}.
For exploit fairness, we use the statistical rate constraint; i.e., constrain the absolute difference between acceptance rates across groups (see {the last paragraph in \cref{sec:algorithm}}). 
When explore fairness is not used -- variants (a), (b) -- the exploration function $g$ is set to be $g^{\rm clf}$
and when explore fairness is used -- variants (c), (d) --
$g$ is set to be $g^{\rm fair}$.
Functions $g^{\rm clf}$, $g^{\rm fair}$ are described in \cref{sec:algorithm}; see \cref{sec:empirical_other} for details of sampling using these functions.

\smallskip
\noindent
\textbf{Baselines.} We compare our approach
to the following baselines:
    (a) \textsc{Kilbertus et al} \cite{kilbertus2020fair}, which uses stochastic classifiers to assign a non-zero exploration probability to every sample;
    (b) \textsc{Yang et al} \cite{yangadaptive}, which employs a bandit-type approach, first determining the likelihoods using a logistic model 
    and then adjusting classifier thresholds to incorporate gathered information;
    (c) \textsc{Rateike et al} \cite{rateike2022don}, which learns an unbiased representation of the data using which an online decision-making model is trained;
    (d) \textsc{Opt-offline}, i.e. the ideal (\textit{unattainable} in partial feedback setting) classifier trained using i.i.d.\ samples from $\mu$;
    (e) \textsc{Fair-clf}, which implements a classifier, with statistical parity and FDR constraints, that is trained every iteration using the available labeled data.
    Implementation details are provided in Appendix~\ref{sec:empirical_other}.
We report the mean and standard error of revenue, FDR, statistical rate, and TPR disparity across protected attributes.
Iteration-wise (i.e., for each $t$, evaluate $f_t$ over $S_{t+1}$ using the above metrics) and cumulative performances are reported to determine short-term and long-term utilities.
\begin{figure*}[t]
    \centering
    \includegraphics[width=0.99\linewidth]{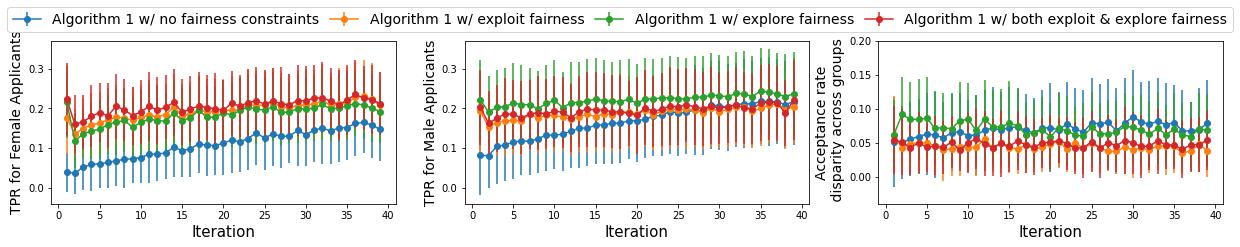} 
    \caption{\small Iteration-wise performance of Algorithm~\ref{alg:main_algorithm} (with explore and/or exploit fairness) on German (protected attribute is gender). 
    Parameters $\alpha{=}0.15$ with $\alpha_{\textrm{exploit}} {=} 0.075 {\cdot} t^{0.2}$ and $\alpha_{\textrm{explore}} {=} \alpha {-} \alpha_{\textrm{exploit}}$, $\tau{=}0.5, \lambda {=} 0, \epsilon{=}10^{-3}$.
    }
    \label{fig:results_explore_exploit_only_ours_german} 
\end{figure*}

\smallskip\noindent
\textbf{Results.} 
Table~\ref{tab:overall_adult_race} presents the cumulative performance of our algorithms for the Adult dataset.
The table shows that all variants of Algorithm~\ref{alg:main_algorithm} achieve high average cumulative revenue while satisfying the given FDR constraint.
Fairness constraints also have an impact on revenue and fairness.
Using either explore or exploit fairness leads to an increase in cumulative revenue. This is because both constraints increase the selection of qualified minority group individuals,
which leads to improved revenue. 
In terms of fairness, \cref{alg:main_algorithm} achieves a low statistical rate for all variants.
Additional fairness is not necessary to achieve a small statistical rate here because the minority group 
form only 10\% of the dataset. Hence, any amount of non-trivial exploration can improve the selection rate for this group.
However, using both explore and exploit fairness leads to the lowest average TPR disparity, showing that additional fairness can be useful in gathering accurate information. 

Figure~\ref{fig:results_explore_exploit_race_only_ours} 
presents the iteration-wise performance of our algorithms.
The first two plots show that 
 TPR increases with increasing iterations for all groups.
TPR is also larger in the initial iterations when using exploit fairness, but is overtaken by or is similar to the TPR of \cref{alg:main_algorithm} with no fairness constraints in the later iterations.
Hence, fairness constraints accelerate data collection in the initial iterations, but once sufficient data is available, it seems to have a similar TPR as the variant with no constraints.
The third plot in Figure~\ref{fig:results_explore_exploit_race_only_ours} also shows that using exploit fairness results in the smallest statistical rate.
With gender as the protected attribute, fairness constraints have a larger impact in reducing these outcome disparities (see results in \cref{sec:empirical_other}).

\smallskip\noindent
\textbf{Comparison to baselines.} 
From Table~\ref{tab:overall_adult_race}, we can see that all variants of Algorithm~\ref{alg:main_algorithm} have slightly smaller revenue than \textsc{opt-offline} baseline.
This is expected since \textsc{opt-offline} is trained using samples from the dataset distribution, which is unavailable in the online setting we operate in. 
When using fairness constraints, our algorithms also achieve lower statistical rates and TPR disparities than \textsc{opt-offline}, addressing the biases in the underlying dataset.
Algorithm~\ref{alg:main_algorithm} also outperforms the \textsc{kilbertus et al} and  \textsc{yang et al} in terms of revenue, statistical rate, and TPR disparity. 
In particular, the revenue for these methods is low because of the high number of false positive errors.
For \textsc{Rateike et al}, FDR, statistical rate, and TPR disparity are small, however, the cumulative revenue achieved is also quite low.
This is because their algorithm results in a large number of false positives in initial iterations and, hence, extremely low revenue in those iterations.
By using FDR constraints, we ensure that the number of false positive errors is small in every iteration.
Algorithm~\ref{alg:main_algorithm} also has better revenue than the \textsc{Fair-clf} baseline, 
which, due to lack of any explicit exploration, gathers outcome information 
at a slower rate than our framework.
For additional assessments of our approach, Appendix~\ref{sec:empirical_other} presents an iteration-wise comparison to \textsc{opt-offline}, \textsc{Fair-clf}, and \textsc{kilbertus et al}.

\subsection{German Credit Dataset}
Next, we evaluate the performance over the German Credit dataset \cite{germanData} which
contains entries of individuals who have taken credit and the task is to decide whether credit risk associated with them is \textit{good} or \textit{bad}.
We use gender as the protected attribute (69\% men, 31\% women).
{Features and pre-processing details are provided in Appendix~\ref{sec:empirical_other}.}
The original dataset contains only 1000 entries. Since this is not sufficient for testing our framework, we sample with replacement 500 entries from the dataset to serve as input $S_t$ for iteration $t$.
Other implementation details are similar to  Section~\ref{sec:adult}.
The utility is measured using $\textrm{revenue}_{c_1, c_2}(\cdot)$  with $c_1 = -500$ and $c_2 = 200$.

\smallskip\noindent
\textbf{Results.} 
Table~\ref{tab:overall_adult_race} presents cumulative performance.
Algorithm~\ref{alg:main_algorithm} with only explore fairness and Algorithm~\ref{alg:main_algorithm} with both explore and exploit fairness constraints achieve higher average revenue than other variants.
Algorithm~\ref{alg:main_algorithm} with only exploit fairness constraints 
also achieves the lowest statistical rate and TPR disparity but the fairness of all variants is within one standard deviation of each other.
Hence, using fairness constraints 
ensures both low disparity and high revenue.
Figure~\ref{fig:results_explore_exploit_only_ours_german} presents iteration-wise performance
and 
shows that average TPR increases as we gather more information. 
Using fairness constraints also leads to high TPR for both groups implying that they accelerate data collection.
Statistical rate is smallest when exploit fairness is used.

\smallskip\noindent
\textbf{Comparison to baselines.} Table~\ref{tab:overall_adult_race} further shows that cumulative revenue using Algorithm~\ref{alg:main_algorithm} is smaller than that of \textsc{opt-offline} and \textsc{kilbertus et al} baselines.
However, these algorithms perform worse in terms of fairness and FDR and lead to relatively larger statistical rates and TPR disparities.
\cref{alg:main_algorithm} with explore fairness or with both fairness constraints also achieve higher revenue and similar fairness as \textsc{Fair-clf} and \textsc{rateike et al}.
Overall, the differences between our algorithm and baselines are relatively smaller for this dataset (compared to Adult) since the dataset size is much smaller here.
Yet, we still see certain improvements due to exploration.

\section{Proofs} \label{sec:proofs}

    \subsection{Proof of \cref{thm:feasibility}}\label{sec:proofof:thm:f
    easibility}
        In this section, we prove \cref{thm:feasibility}.
        For ease of reference, we restate \cref{thm:feasibility} below.

        \thmFeasibility*

        Recall that we assume that \cref{assump:f0_is_feasible} holds with constants $\alpha,\lambda\in (0,1]$.
        This implies that the classifier $f_0\in \cF$ satisfies the $\alpha$-FDR constraint and has a selection rate of at least $\lambda$, i.e., 
        \[
            \frac{\Pr_\mu\insquare{f_0(X,Z)=1 \text{ and } Y=0}}{\Pr_\mu\insquare{f_0(X,Z) = 1}}\leq \alpha
            \quad\text{and}\quad 
            \Pr_\mu\insquare{f_0(X,Z)=1}\geq \lambda.
            \yesnum\label{eq:prop_f0}
        \]
        Where we use $\Pr_\mu[\cdot]$ to denote $\Pr_{(X,Y,Z)\sim \mu}[\cdot]$ and $\Ex_\mu[\cdot]$ to denote $\Ex_{(X,Y,Z)\sim \mu}[\cdot]$. 
        
        To prove \cref{thm:feasibility}, given any $\eps,\delta,\tau\in (0,1]$, we need to show that if $n$ is sufficiently large, 
        then at each iteration $t\in \inbrace{1,2,\dots}$, with probability at least $1-\delta$, the data collection and prediction framework in \cref{alg:main_algorithm},
        run with the parameters $\alpha_{\rm exploit}$ and $\alpha_{\rm explore}$,
        makes predictions that satisfy the $\inparen{\alpha_{\rm exploit}+\alpha_{\rm explore}+O(\eps)}$-FDR constraint, i.e., %
        \[
            \frac{
                \sum_{(x,z)\in S_t}\mathbb{I}\insquare{\hy_{(x,z)} = 1 \text{ and } y = 0}
            }{
                \sum_{(x,z)\in S_t}\mathbb{I}\insquare{\hy_{(x,z)} = 1}
            }
            \leq \alpha_{\rm exploit}+\alpha_{\rm explore}+O\inparen{\eps}.
            \yesnum\label{eq:required_fdr_const}
        \]
        In particular, we will show that the following lower bound on $n$ is sufficient 
        \[
            n
            \geq 
            \frac{12\abs{D}}{\tau \eps^3}\cdot \frac{1}{\alpha_{\rm explore} \lambda} \cdot \log{\frac{\abs{D}}{\delta\sigma}}.
            \yesnum\label{eq:lb_on_n}
        \]

        \renewcommand*{\arraystretch}{1.5}

        \noindent The main step in the proof is to establish the following concentration inequality.
        \begin{lemma}\label{lem:conc_inequality}
            For any bounded function $h\colon \zo\times\zo\times [p]\to [0,1]$,
            any number $t\in \inbrace{1,2,\dots}$,
            and any constants $\eps,\tau,\delta\in (0,1]$, 
            given 
            $n \geq \abs{D}\cdot \poly\inparen{\nfrac{1}{\tau},\nfrac{1}{\eps},\nfrac{1}{\lambda},\nfrac{1}{\alpha_{\rm explore}}}  \cdot \log\inparen{\nfrac{\abs{D}}{(\delta\sigma)}}$
            at the $t$-th iteration $\eta_w$ is such that the following holds
            \[
                \Pr\insquare{
                    \forall_{f\in \cF},\quad 
                    \abs{
                        \begin{array}{l}
                            \ \ \Ex_{\eta_w}\insquare{h(f(X,Z),Y,Z) \mid (X,Z)\in \expl_t }\\
                            - \ \ 
                            \Ex_{\mu}\insquare{h(f(X,Z),Y,Z)\mid (X,Z)\in \expl_t }
                        \end{array}
                    }
                    \leq \eps
                }
                \geq 1-\delta.
            \]
            Where $\eta_w$ is the distribution defined in Step~3 of \cref{alg:main_algorithm}, $\expl_t$ is the exploitation region at the $t$-th iteration. 
            The expectations and probabilities are over the randomness in the draw of $S_0,S_1,\dots,S_t$ and the randomness in \cref{alg:main_algorithm}.
        \end{lemma}
        Before presenting the proof of \cref{thm:feasibility} we require the definition of the Vapnik–Chervonenkis (VC) dimension.
        \begin{definition}
        	Given a finite set $A$, define the collection of subsets $\cF_A\coloneqq \{\{a\in A\mid f(a)=1\}\mid f\in \cF\}$.
        	We say that $\cF$ shatters a set $B$ if $|\cF_{B}|=2^{|B|}$.
        	The VC dimension of $\cF$, ${\rm VC}(\cF)\in \N$,  {is the largest integer such that there exists a set $C$ of size ${\rm VC}(\cF)$ that is shattered by $\cF$.}
        \end{definition}
        \begin{proof}[Proof of \cref{thm:feasibility} assuming \cref{lem:conc_inequality}]
            Fix any iteration $t\in \inbrace{1,2,\dots}$.
            Assume $\eps\leq \sfrac{\lambda}{3}$.\footnote{If $\eps> \sfrac{\lambda}{3}$, we can set $\eps=\sfrac{\lambda}{3}$. This only improves the guarantee we prove, and the lower bound on $n$ is not violated as it depends on $\min\inbrace{\sfrac{\lambda}{3},\eps}$.}
            Consider the classifier $f_t$ in \cref{alg:main_algorithm}.
            Recall that \cref{alg:main_algorithm} uses $f_t$ for making predictions in the exploitation region $\expl_t$ and makes at most $\phi$ positive predictions in the  exploration region $\expr_t$, where
            \[
                \phi \coloneqq \inparen{
                    \sum_{(x,z)\in S_t\cap \expl_t}\mathbb{I}\insquare{f_t(x,z) = 1}
                }\cdot \alpha_{\rm explore}.
            \]
            Due to this, to establish \cref{eq:required_fdr_const}, it suffices to prove the following
            \[
                \frac{
                    \sum_{(x,z)\in S_t\cap \expl_t}\mathbb{I}\insquare{f_t(x,z) = 1\text{ and } y = 0}
                }{
                    \sum_{(x,z)\in S_t\cap \expl_t}\mathbb{I}\insquare{f_t(x,z) = 1}
                }
                \leq 
                \alpha_{\rm exploit}+\frac{10\eps}{\lambda}.
                \yesnum\label{eq:toprove}
            \]
            First, we will express the above ration as a ratio of expectations over $\mu$ using a standard generalization bound.
            Then, we will use \cref{lem:conc_inequality} to complete the proof.
            Note that, we can bound the VC-dimension of $\cF$ by $\abs{D}$: ${\rm VC}(\cF)\leq \abs{D}$ \cite{shalev2014understanding}.
            Hence, using the standard generalization inequality in Section 28.1 \cite{shalev2014understanding} and the lower bound on $n$, we have the following bound:
            for any distribution $\zeta$ over $D$
            \[
                \Pr\insquare{
                    \forall_{f\in \cF},\quad 
                    \abs{
                        \Ex_{S\sim \zeta^n}\insquare{h(f(X,Z),Y,Z) }
                        - 
                        \Ex_{\zeta}\insquare{h(f(X,Z),Y,Z)}
                    }
                    \leq \eps
                }
                \geq 1-\delta.
            \]
            Setting $\zeta$ to be the distribution $\mu$ restricted to the $\expl_t$ and observing that $S_t$ has $n$ samples drawn i.i.d.\ from $\mu$, 
            we deduce the following from the above inequality
            \[
                \Pr\insquare{
                    \forall_{f\in \cF},\quad 
                    \abs{
                        \begin{array}{l}
                            \ \ \Ex_{S_t}\insquare{h(f(X,Z),Y,Z) 
                        \mid (X,Z)\in \expl_t}\\
                            - \ \ 
                            \Ex_{\mu}\insquare{h(f(X,Z),Y,Z)
                        \mid (X,Z)\in \expl_t}
                        \end{array}
                    }
                    \leq \eps
                }
                \geq 1-\delta.\yesnum\label{eq:conc1}
            \]
            Combining \cref{lem:conc_inequality}, \cref{eq:conc1} using the triangle inequality, and the union bound, we get the following bound 
            \[
                \Pr\insquare{
                    \forall_{f\in \cF},\quad 
                    \abs{
                        \begin{array}{l}
                            \ \ \Ex_{\eta_w}\insquare{h(f(X,Z),Y,Z) 
                        \mid (X,Z)\in \expl_t}\\
                            - \ \ 
                            \Ex_{S_t}\insquare{h(f(X,Z),Y,Z)
                        \mid (X,Z)\in \expl_t}
                        \end{array}
                    }
                    \leq 2\eps
                }
                \geq 1-2\delta.\yesnum\label{eq:conc2}
            \]
            Let $\evE$ be the event that the following holds
            \[
                \forall_{f\in \cF},\quad 
                    \abs{
                        \begin{array}{l}
                            \ \ \Ex_{\eta_w}\insquare{h(f(X,Z),Y,Z) 
                        \mid (X,Z)\in \expl_t}\\
                            - \ \ 
                            \Ex_{S_t}\insquare{h(f(X,Z),Y,Z)
                        \mid (X,Z)\in \expl_t}
                        \end{array}
                    }
                    \leq 2\eps.
            \]
            The \cref{eq:conc2} implies that $\Pr[\evE]\geq 1-2\delta$.
            Conditioned on $\evE$, selecting 
            \[
                h(y_1,y_2,z)=\mathbb{I}[y_1=1\text{ and }y_2=0],
            \]
            \cref{eq:conc1} implies that
            \begin{align*}
                &
                    \abs{
                        \begin{array}{l}
                            \ \Ex_{\eta_w}\insquare{
                    f_t(X,Z)= 1\text{ and } Y=0
                        \mid (X,Z)\in \expl_t}\\
                            -\qquad 
                            \frac{\sum\nolimits_{(x,z)\in S_t\cap \expl_t}\mathbb{I}\insquare{f_t(x,z) = 1 \text{ and } y = 0}}
                            {
                                \abs{S_t\cap \expl_t}
                            }
                        \end{array}
                    }
                    \leq 2\eps.
                    \yesnum\label{eq:bound1}
            \end{align*}
            Similarly, conditioned on $\evE$, for 
            \[
                h(y_1,y_2,z)=\mathbb{I}[y_1=1],
            \] 
            \cref{eq:conc1} implies that
            \[
                \abs{
                    \Ex_{\eta_w}\insquare{f_t(X,Z)= 1
                        \mid (X,Z)\in \expl_t}
                    -
                    \frac{\sum\nolimits_{(x,z)\in S_t\cap \expl_t}\mathbb{I}\insquare{f_t(x,z) = 1}}
                    {
                        \abs{S_t\cap \expl_t}
                    }
                }
                \leq 
                    2\eps.
                    \yesnum\label{eq:bound2}
            \]
            Since $f_t$ is feasible for \prog{prog:main_algorithm} in Step~3 of \cref{alg:main_algorithm}, it follows that $f_t$ satisfies the following constraints
            \begin{align*}
                \frac{
                    \Pr_{\eta_w}\insquare{f_t(X,Z) = 1 \textrm{ and } Y=0}
                }{
                    \Pr_{\eta_w}\insquare{f_t(X,Z)=1}
                }
                    \leq \alpha_{\rm exploit}+\eps
                \quad\textrm{and}\quad
                \Pr_{\eta_w}\insquare{f_t(X,Z)=1}
                \geq \lambda - \eps.
                \yesnum\label{eq:guarantee_f0}
            \end{align*}
            Now, we are ready to prove \cref{eq:toprove}.
            \begin{align*}
                &\frac{
                    \sum_{(x,z)\in S_t\cap \expl_t}\mathbb{I}\insquare{
                        f_t(x,z) =1 \textrm{ and } \calO(x,z)=0
                    }
                }{
                    \sum_{(x,z)\in S_t\cap \expl_t}\mathbb{I}\insquare{f_t(x,z) = 1}
                }\\
                &\qquad \Stackrel{\eqref{eq:bound1},\ \eqref{eq:bound2}}{\leq} \quad 
                    \frac{
                        \Pr_{\eta_w}\insquare{
                            f_t(X,Z) =1 \textrm{ and } Y=0
                        } + 2\eps
                    }{
                        \Pr_{\eta_w}\insquare{f_t(X,Z)=1} - 2\eps
                    }\\
                &\qquad \Stackrel{}{=} \quad 
                    \frac{
                        \Pr_{\eta_w}\insquare{
                            f_t(X,Z) =1 \textrm{ and } Y=0
                        } + 2\eps
                    }{
                        \Pr_{\eta_w}\insquare{f_t(X,Z)=1}
                    }
                    \cdot \frac{1}{1-\frac{2\eps}{\Pr_{\eta_w}\insquare{f_t(X,Z)=1}}}\\
                &\qquad \Stackrel{\eqref{eq:guarantee_f0}}{\leq} \quad 
                    \frac{
                        \Pr_{\eta_w}\insquare{
                            f_t(X,Z) =1 \textrm{ and } Y=0
                        } + 2\eps
                    }{
                        \Pr_{\eta_w}\insquare{f_t(X,Z)=1}
                    }
                    \cdot \frac{1}{1-\frac{2\eps}{\lambda-2\eps}}\\
                &\qquad \leq  \quad 
                    \frac{
                        \Pr_{\eta_w}\insquare{
                            f_t(X,Z) =1 \textrm{ and } Y=0
                        } + \eps
                    }{
                        \Pr_{\eta_w}\insquare{f_t(X,Z)=1}
                    }
                    \cdot \inparen{1+\frac{4\eps}{\lambda}}.
                    \customlabel{eq:penultimate}{\theequation}
                    \tagnum{Using that $\inparen{1-\frac{x}{1-2x}}^{-1}\leq 1+4x$ for all $x\in [0,\frac{1}{4}]$ and $\eps\leq \frac{\lambda}{4}$}
            \end{align*}
            An upper bound on the RHS of the above equation is as follows 
            \begin{align*}
                &\frac{
                        \Pr_{\eta_w}\insquare{
                            f_t(X,Z) =1 \textrm{ and } Y=0
                        } + 2\eps
                    }{
                        \Pr_{\eta_w}\insquare{f_t(X,Z)=1}
                    }
                    \cdot \inparen{1+\frac{4\eps}{\lambda}}\\
                &\qquad \Stackrel{\eqref{eq:guarantee_f0}}{\leq} \quad
                    \frac{
                            \Pr_{\eta_w}\insquare{
                                f_t(X,Z) =1 \textrm{ and } Y=0
                            }
                        }{
                            \Pr_{\eta_w}\insquare{f_t(X,Z)=1}
                        }
                        \cdot \inparen{1+\frac{4\eps}{\lambda}}
                    + \frac{2\eps}{\lambda-\eps}
                    \cdot \inparen{1+\frac{4\eps}{\lambda}}\\
                &\qquad \leq  \quad
                    \frac{
                            \Pr_{\eta_w}\insquare{
                                f_t(X,Z) =1 \textrm{ and } Y=0
                            }
                        }{
                            \Pr_{\eta_w}\insquare{f_t(X,Z)=1}
                        }
                        \cdot \inparen{1+\frac{4\eps}{\lambda}}
                    + \frac{4\eps}{\lambda}
                    \cdot \inparen{1+\frac{4\eps}{\lambda}}
                    \tag{Using that $\frac{x}{1-x}\leq 2x$ for all $x\in [0,\frac{1}{4}]$ and $\eps\leq \frac{\lambda}{4}$}\\
                &\qquad \Stackrel{\eqref{eq:guarantee_f0}}{\leq} \quad
                    (\alpha_{\rm exploit}+\eps)\cdot \inparen{1+\frac{4\eps}{\lambda}}
                    + \frac{2\eps}{\lambda}
                    \cdot \inparen{1+\frac{4\eps}{\lambda}}\\
                &\qquad \leq\quad \alpha_{\rm exploit} + \frac{14\eps}{\lambda}.
                \tag{Using that $4\eps \leq \lambda$, $\alpha\leq 1$, and $0\leq \lambda \leq 1$}
            \end{align*}
            Substituting this in Equation~\eqref{eq:penultimate}, implies \cref{eq:toprove}.
            The result follows as $\evE$ holds with probability at least $1-2\delta$.
        \end{proof}
        \begin{remark}
            Note that the above proof does not use \cref{assump:f0_is_feasible}.
            This assumption is only required to ensure that \prog{prog:main_algorithm} in Step~3 of \cref{alg:main_algorithm} is feasible for all $t\in \inbrace{1,2,\dots}$.
        \end{remark}
        \renewcommand*{\arraystretch}{1}
        
        \subsubsection{Proof of \cref{lem:conc_inequality}}
            \label{sec:discussion_of_techniques}
        \paragraph{Discussion of techniques.}
        The generalization bound in \cref{lem:conc_inequality} shows that the reweighted distribution $\eta_w$ in Step~3 of \cref{alg:main_algorithm} is a good approximation of $\mu$ on $\expl_t$.
        Recall that this bound is as follows:
        if $n\geq n_0\coloneqq \abs{D}\cdot\poly\big(\nfrac{1}{\lambda},\nfrac{1}{\tau}, \nfrac{1}{\min\inbrace{\eps,\alpha-\eps}}\big)\cdot \log\inparen{\nfrac{\abs{D}}{\sigma\delta}}$, then 
        at any iteration $t$ and for any bounded function $h\colon \zo\times \zo\times [p]\to[0,1]$ the following holds
        \[
            \Pr\insquare{
                    \forall_{f\in \cF},\quad 
                    \abs{
                        \begin{array}{l}
                            \ \ \ \ \Ex_{\eta_w\mid (X,Z)\in \expl_t}\insquare{h(f(X,Z),Y,Z)}\\
                            - \ \ 
                            \Ex_{\mu\mid (X,Z)\in \expl_t}\insquare{h(f(X,Z),Y,Z) }
                        \end{array}
                    }
                    \leq O(\eps)
                }
                \geq 1-\delta.
                \yesnum\label{eq:generalizationBound}
        \]
        At any $t$, $\eta_w$ has the following product form
        \[
            \eta_w(x,y,z)  
                \propto \mathds{I}\insquare{(x,z)\in \expl_t} 
                \cdot 
                \Pr\limits_{S_0,\dots,S_t}\insquare{(X,Z)=(x,z)}
                \cdot \Pr\limits_{S_0,\dots,S_t}\insquare{(x,y,z)\in \bigcup_{i=0}^{t-1} L_i}
                \yesnum\label{eq:expEta}
        \]
        The first term ensures that the support of $\eta_w$ is $\expl_t$.
        The second term is an estimate of $\mu(x,z)$. 
        The third term is an estimate of $\Pr_\mu\insquare{Y{=}y\mid (X,Z){=}(x,z)}$.
        If both of these estimates are exact, then \cref{eq:generalizationBound} follows.
        We prove that the estimates are nearly correct for sample $(x,z)$ with a sufficient probability of mass under $\mu$.
        Consider the set $R$ of all samples that have a small probability mass under $\mu$:
        $R\coloneqq \inbrace{(x,z)\mid \mu(x,z)\leq \nfrac{\eps}{\abs{D}}}.$
        The proof of the claim for the first estimate is straightforward:
        the Chernoff bound and the union bound imply that if $n\geq n_0$, then 
        \[
            \Pr\insquare{\forall_{(x,z)\in \expl_t\setminus R},\quad \Pr_{S_t}\insquare{(X,Z)=(x,z)}\in (1\pm \eps)\cdot \mu(x,z) } \geq  1 - \delta. 
            \yesnum\label{eq:conc1a}
        \]
        The claim for the second estimate holds if $\bigcup_{i=0}^{t-1} L_i$ has at least $k_0\coloneqq \poly\inparen{\nfrac{1}{\eps}}\cdot\log\inparen{\nfrac{\abs{D}}{\delta}}$ copies of each $(x,z)\in \expl_t\setminus R$.
        Under this assumption, standard techniques imply that
        \[
            {  \Pr\insquare{ 
                \forall_{(x,z)\in \expl_t\setminus R },\ 
                \text{\small $\Pr\limits_{S_0,\dots,S_t}\insquare{(x,y,z) {\in} \bigcup\limits_{i=0}^{t-1} L_i} 
                \in 
                (1\pm \eps) \Pr\limits_\mu\insquare{Y{=}y\mid (X,Z){=}(x,z)}$}
            } 
            \geq  1 {-} \delta.}  
            \yesnum\label{eq:conc2a}
        \]
        Together, \cref{eq:conc1a,eq:conc2a}, imply that: with probability at least $1-\delta$, 
        $\eta_w(x,y,z)\in (1\pm O(\eps))\cdot \mu(x,y,z)$ for any $(x,z)\in \expl_t\setminus R$ and $y\in \zo$.
        That is, $\eta_w$ is a good approximation of $\mu$ on $\expl_t\setminus R$.
        Since the total probability mass of samples in $R$ is at most $\eps$, the required generalization bound (\cref{eq:generalizationBound}) follows.
        It remains to show that there are at least $k_0$ copies of each $(x,z)\in \expl_t\setminus R$ in $\bigcup_{i=0}^{t-1} L_i$ (which was used to prove \cref{eq:conc2a}).
        
        Fix any $(x,z)\in \expl_t\setminus R$.
        Let $N_i$ be the number of copies of $(x,z)$ in $L_i$ and $N=\sum_{i=0}^{t-1} N_i$ be the number of copies of $(x,z)$ in $\bigcup_{i=0}^{t-1} L_i$. 
        One can show that $\Ex\insquare{N_i}=\Omega(n\times g(x,z; f_i))$. %
        Since $(x,z)\in \expl_t$, $\sum_{i=0}^t g(x,z;f_i)\geq \tau$ and, hence, by linearity of expectation $\Ex\insquare{N}\geq \Omega(n\times \tau) =  k_0$.
        Thus, it suffices to show that $N\geq \Omega\inparen{\Ex\insquare{N}}$ with high probability.
        Here, $N$ is a sum of 0/1 random variables $Z_{ij}$: denoting whether $(x,z)$ is the $j$-th sample in $L_i$.
        If $\inbrace{Z_{ij}}_{i,j}$ are independent, then one may hope to prove the concentration of $N$ via standard inequalities. 
        
        The main challenge is that, since \cref{alg:main_algorithm}'s choices at the $i$-th iteration depend on past observations, independence may not hold.
        Let $T$ be the last iteration when $(x,z)$ is in the exploration region.
        We overcome this using the fact that $\inbrace{Z_{ij}\mid i\in [T],j\in \abs{L_i}}$ are mutually independent because till $(x,z)$ is included in the exploitation region, labels observed for $(x,z)$ do not affect $\eta_w$ and, hence, \cref{alg:main_algorithm}'s choices.

        \noindent\paragraph{Proof of \cref{lem:conc_inequality}.}
            Recall the following definitions 
            \begin{align*}
                \sigma &\coloneqq \min_{z\in [p]} \sigma_z,\quad 
                \text{where},\quad 
                \forall z\in [p],\quad 
                \sigma_z 
                    \coloneqq \min_{f\in \cF} \frac{\min_{x\in \cX} g(x,z; f)}{\sum_{(x,\ell)\in \cD} g(x,\ell;,f)}.
                \yesnum\label{def:sigma}
            \end{align*}
            Fix any 
            bounded function $h\colon \zo\times\zo\times [p]\to [0,1]$,
            any number $t\in \N$, and 
            any constants $\eps,\tau,\delta\in (0,1]$.
            Fix any $n$ satisfying the following lower bound 
            \[
                n \geq \frac{12\abs{D}}{\tau \eps^3}\cdot \frac{1}{\alpha_{\rm explore} \lambda} \cdot \log{\frac{\abs{D}}{\delta\sigma}}.
            \]
            The proof is divided into three steps.
            In the first step, we show that, when $D$ is finite and $n$ is sufficiently large, 
                with probability at least $1-\delta$, at each $t$, 
                it holds that: for each  $(x,z)\in \expl_t$ satisfying $\mu(x,z)\geq \frac{\eps}{\abs{D}}$
            \[
                \Pr_{(X,Y,Z)\sim S_i}[(X,Z)=(x,z)]\in (1\pm\eps)^2 \cdot \mu(x,z).
            \]
            In the second step, we show that with probability at least $1-\delta$, at each $t$, 
                it holds that: for each  $(x,z)\in \expl_t$ satisfying $\mu(x,z)\geq \frac{\eps}{\abs{D}}$ and $y\in \zo$
                \[
                    \Pr_{(X,Y,Z)\sim L_0\cup L_1\cup\dots\cup L_t}\insquare{Y=y\mid (X,Z)=(x,z)}
                    \in (1\pm \eps)^2 
                    \cdot 
                    \Pr_{(X,Y,Z)\sim \mu}\insquare{Y=y\mid (X,Z)=(x,z)}.
                \]
            In the third step, we conclude the proof. %

        \paragraph{Step 1 ($\Pr_{S_i}[(X,Z)=(x,z)]\in (1\pm\eps)^2 \cdot \mu(x,z)$ for each $(x,z)\in \expl_t$):}
            Fix any $(x,z)\in D$ with $\mu(x,z)\geq \frac{\eps}{\abs{D}}$.
            Since for each $t$, $S_t$ contains of $n$ iid samples from $\mu$, it follows that the expected number of copies of $(x,z)$ in $S_t$, say $N_{x,z,t}$, is 
            \[
                \Ex\insquare{N_{x,z,t} }
                = n\cdot \mu(x,z)
                \qquad \Stackrel{\mu(x,z)\geq \frac{\eps}{\abs{D}}}{\geq} \qquad 
                \frac{12}{\tau \eps^2}\cdot \frac{1}{\alpha_{\rm explore} \lambda } \cdot \log{\frac{\abs{D}}{\delta\sigma}}.
                \yesnum\label{eq:meanLB1}
            \]
            Moreover, the Chernoff bound implies that with a probability of at least 
            \begin{align*}
                1-2\exp\inparen{-\frac{\eps^2}{3}\cdot \frac{12}{\tau \eps^2}\frac{1}{\alpha_{\rm explore} \lambda } \cdot \log{\frac{\abs{D}}{\delta\sigma}}}
                & 
                \geq 1-2\frac{\delta\sigma}{\abs{D}} 
                \tagnum{Using that $0\leq \tau\alpha_{\rm explore}\lambda \leq 1$}
                \customlabel{eq:whpEvent1}{\theequation}
            \end{align*} 
            $N_{x,z,t}$ lies in the following interval 
            \[
                \insquare{(1-\eps)n\cdot \mu(x,z), (1+\eps)n\cdot \mu(x,z)}.
                \yesnum\label{eq:whpEvent2}
            \]
            Now the union bound over all $x\in \cD$ implies that with a probability of at least $1 - 2\delta\sigma,$ the following holds:
            for all $(x,z)\in \expl_t$ satisfying $\mu(x,z)\geq \frac{\eps}{\abs{D}}$
            \[
                \Pr_{S_t}[(X,Z)=(x,z)]
                = \frac{N_{x,z,t}}{n}
                \stackrel{\eqref{eq:whpEvent1},\eqref{eq:whpEvent2}}{\in}
                (1\pm\eps)^2 \cdot \mu(x,z).
                \yesnum\label{eq:Step1ConcBound}
            \]

        \paragraph{Step 2 ($\Pr_{L_0\cup\dots\cup L_t}\insquare{y\mid (x,z)} \in (1\pm \eps)^2 \cdot \Pr_{\mu}\insquare{y\mid (x,z)}$ for each $(x,z)\in \expl_t$):}
            Fix any $(x,z)\in D$ with $\mu(x,z)\geq \frac{\eps}{\abs{D}}$.
            Let $T_{x,z}\in \N$ be the last iteration where $(x,z)$ is in the exploration region.
            Since $g(x,z;f_i)\geq \sigma$ for each $i\in \N$, it follows that 
            \[
                T_{x,z} \leq \sigma^{-1}.
                \yesnum\label{eq:UBOnT}
            \]
            For any $i\in \N$, let $N_{x,z,i}$ be the number of copies of $(x,z)$ in $S_i$.
            Let $\evE_{x,z}$ be the event that 
            \[
                \forall 1\leq i\leq T_{x,z},\quad N_{x,z,i} \geq \frac{6}{\tau \eps^2}\cdot \frac{1}{\alpha_{\rm explore} \lambda } \cdot \log{\frac{\abs{D}}{\delta\sigma}}.
            \]
            The Chernoff bound and the union bound imply that 
            \[
                \Pr\insquare{\evE_{x,z}} \geq  1-T_{x,z}\times 2\frac{\delta\sigma }{\abs{D}} 
                \stackrel{\eqref{eq:UBOnT}}{\geq} 1-2\frac{\delta}{\abs{D}}.
            \]
            Order the $N_{x,z,i}$ copies of $(x,z)$ in $S_i$ arbitrarily for each $i\in \N$.
            Observe that each copy of $(x,z)$ in $S_i$ (for any $i\in [T_{x,z}]$) is positively labeled with probability $\alpha_{\rm explore} \lambda \cdot g(x,z; f_i)$.
            Henceforth, we abbreviate $g(x,z; f_i)$ as $g_i(x,z)$.
            The event that $(x,z)$ is positively labeled in iterations $1\leq i_1,i_2\leq T_{x,z}$ are independent as till $(x,z)$ is not in the exploitation region, it does not affect \cref{alg:main_algorithm}'s decision to positively label samples.
            Let $Z_j\in \zo$ be the indicator random variable that the $j$-th copy of $(x,z)$ in $S_i$ exists and is positively labeled for some $1\leq i\leq T_{x,z}$.
            Define 
            \[
                \Delta\coloneqq \frac{6}{\tau \eps^2}\cdot \frac{1}{\alpha_{\rm explore} \lambda } \cdot \log{\frac{\abs{D}}{\delta\sigma}}.
                \yesnum\label{def:DeltaTmp}
            \]
            Conditioned on $\evE_{x,z}$ it holds that:
            for all $1\leq j\leq \Delta$
            \begin{align*}
                \Pr\insquare{Z_j=1 \mid \evE_{x,z}}
                &= 1 - \prod_{i=0}^T \inparen{ 1 - g_i(x,z) \alpha_{\rm explore}\lambda}\\
                &\geq 1 - \prod_{i=0}^T \exp\inparen{- g_i(x,z) \alpha_{\rm explore}\lambda} \tag{Using that $e^{-x}\geq 1-x$ for all $x\in \R$}\\
                &= 1 -\exp\inparen{-\sum_{i=0}^T g_i(x,z) \alpha_{\rm explore}\lambda} \\
                &\geq 1 -\exp\inparen{-\tau\alpha_{\rm explore}\lambda} \tag{Using that $T$ is the last iteration when $(x,z)$ is in the exploration region and, hence, $\sum_{i=0}^T g_i(x,z)\geq \tau$}\\
                &\geq 1 -\inparen{1-\frac{1}{2}\tau\alpha_{\rm explore}\lambda}  \tag{Using that $e^{-x}\leq 1-\frac{x}{2}$ for all $x\in [0,1]$ and $0\leq \tau\alpha_{\rm explore}\lambda\leq 1$}\\
                &= \frac{1}{2}\tau\alpha_{\rm explore}\lambda.
                \yesnum\label{eq:LBOnProbOfW}
            \end{align*}
            Moreover, $Z_i$ and $Z_j$ are independent for all $i\neq j$.
            Hence, the Chernoff bound implies that 
            \begin{align*}
                \Pr\insquare{\frac{\sum_{j=0}^{\Delta} Z_j}{\Delta} \geq \frac{1}{2}\cdot \frac{1}{2}\tau\alpha_{\rm explore}\lambda}
                \ \ &\geq\ \  1-2\exp\inparen{ -\frac{1}{3}\cdot \frac{1}{4}\cdot \frac{\Delta}{2}\tau\alpha_{\rm explore}\lambda}\\ 
                \ \ &\Stackrel{\eqref{def:DeltaTmp}}{=}\ \  1-2\exp\inparen{ -\frac{\tau\alpha_{\rm explore}\lambda}{4\tau\eps^{-2}\alpha_{\rm explore}\lambda}\cdot \log{\frac{\abs{D}}{\delta \sigma}}}\\ 
                \ \ &\Stackrel{\eqref{def:DeltaTmp}}{=}\ \  1-2\frac{\delta \sigma}{\abs{D}}.
                \tagnum{Using $0<\eps\leq \frac{1}{2}$}
                \customlabel{eq:ConcSumW}{\theequation}
            \end{align*}
            Let $P_{x,z,i}$ be the number of copies of $(x,z)$ that are positively labeled in the first $i$ iterations.
            Observe that $P_{x,z,\Delta}\geq \sum_{j=0}^{\Delta} Z_j$.
            Thus, the union bound implies that with probability at least $1-2\delta$, for each $(x,z)\in \expl_t$ satisfying $\mu(x,z)\geq \frac{\eps}{\abs{D}}$, at least
            \[
                P_{x,z,t}\geq P_{x,z,\Delta}\geq \frac{3}{2} \eps^{-2} \log{\frac{\abs{D}}{\delta \sigma}}
            \]
            copies of $(x,z)$ are positively labeled in the first $t$ iterations. 
            Since each positively lebled copy of $(x,z)$ is inclded in $\bigcup_{i=0}^{t-1} L_i$, it follows that with probability at least $1-2\delta$, at least $\frac{3}{2\eps^{2}} \log{\frac{\abs{D}}{\delta \sigma}}$ copies of $(x,z)$ are included in $\bigcup_{i=0}^{t-1} L_i$ for each $(x,z)\in \expl_t$ satisfying $\mu(x,z)\geq \frac{\eps}{\abs{D}}$.
            Suppose this event is $\evE_t$.
            Consider $(X,Y,Z)\sim \mu$.
            Since conditioned on $(X,Z)=(x,z)$, $Y$ is a bernoulli random variable with 
            \[
                \Pr[Y=1]=\Pr_\mu[Y=1\mid (X,Z)=(x,z)],   
                \yesnum\label{eq:bernoulli}
            \]
            conditioned on $\evE_t$, it holds that 
            \begin{align*}
                \Pr_{(X,Y,Z)\sim L_0\cup L_1\cup\dots\cup L_t}\insquare{Y=y\mid (X,Z)=(x,z)}
                &= \frac{\sum_{(X,Y,Z)\in L_0\cup\dots\cup L_t\mid (X,Z)=(x,z)}\mathbb{I}\insquare{Y=y}}{P_{x,z,t}}.%
            \end{align*}
            Using standard concentration properties of sums of Bernoulli random variables and the fact that conditioned on $\evE_t$, $P_{x,z,t}\geq \frac{3}{2\eps^{2}} \log{\frac{\abs{D}}{\delta \sigma}}$, it follows, that with probability at least $1-\frac{\delta\sigma}{\abs{D}}$
            \[
                \sum_{(X,Y,Z)\in L_0\cup\dots\cup L_t\mid (X,Z)=(x,z)}\mathbb{I}\insquare{Y=y}
                \stackrel{\eqref{eq:bernoulli}}{\in} (1\pm\eps) \cdot P_{x,z,t}\cdot \Pr_\mu[Y=1\mid (X,Z)=(x,z)].
            \]
            Thus, conditioned  on event $\evE_t$, with probability at least $1-\frac{\delta\sigma}{\abs{D}}$, it holds that 
            \[
                \Pr_{(X,Y,Z)\sim L_0\cup L_1\cup\dots\cup L_t}\insquare{Y=y\mid (X,Z)=(x,z)}
                \in (1\pm \eps) \Pr_\mu[Y=1\mid (X,Z)=(x,z)].
            \]
             The union bound now implies that, conditioned on $\evE_t$, with probability at least $1-2\delta\sigma$, for an $(x,z)\in \expl_t$ satisfying $\mu(x,z)\geq \frac{\eps}{\abs{D}}$, it holds that 
            \[
                \Pr_{(X,Y,Z)\sim L_0\cup L_1\cup\dots\cup L_t}\insquare{Y=y\mid (X,Z)=(x,z)}
                \in (1\pm \eps) \Pr_\mu[Y=1\mid (X,Z)=(x,z)].
            \]
            This completes Step 2.

        \noindent\paragraph{Step 3 (Completing the proof of \cref{lem:conc_inequality}): }
            Let $R$ be the set of all samples $(x,z)$ such that $\mu(x,z)\leq \frac{\eps}{\abs{D}}$:
            \[
                R\coloneqq \inbrace{ (x,z)\in \cX\times [p]\mid \mu(x,z)\leq \frac{\eps}{p \abs{D}}}.
            \]
            Since $\abs{S}\leq \abs{D}$, $\mu(S)\leq \eps.$
            Hence, the Chernoff bound implies that with probability at least $1-\delta$,
            \[
                \abs{R\cap S_t}\leq n\eps + \sqrt{n\eps\log{\frac{1}{\delta}}}.
                \yesnum\label{eq:QuickChernoffBound}
            \]
            Consequently, with probability at least $1-\delta$, it holds that
            \begin{align*}
                \Pr_{S_t}[(X,Z)\in R]
                &\stackrel{\eqref{eq:QuickChernoffBound}}{\leq} 
                \frac{n\eps + \sqrt{n\eps\log{\frac{\abs{D}}{\delta} }}}{n}\\
                &\leq \eps + \eps^2 \sqrt{
                    \frac{\tau \alpha_{\rm explore}\lambda}{12\abs{D}}
                }\tag{Using that $n\geq \frac{12\abs{D}}{\tau \eps^3}\cdot \frac{1}{\alpha_{\rm explore} \lambda} \cdot \log{\frac{\abs{D}}{\delta\sigma}}$}\\
                &\leq 2\eps. \tag{Using that $\abs{D}\geq 1$, $0\leq \alpha\lambda\tau \leq 1$, and $\eps\leq 1$}
            \end{align*}
            For any $t$, let $\evG_t$ be the event that the following holds:
            \begin{align*}
                \Pr_{S_t}[(X,Z)\in R] &\leq 2\eps, 
                \yesnum\label{eq:bound133}\\ 
                \forall (x,z)\in \expl_t\setminus R,
                \qquad \qquad \qquad 
                \Pr_{S_t}[(X,Z)=(x,z)]
                &\in
                (1\pm\eps)^2 \cdot \mu(x,z),\\
                \forall (x,z)\in \expl_t\setminus R,\ \ 
                \Pr_{L_0\cup\dots\cup L_t}\insquare{Y{=}y\mid (X,Z){=}(x,z)}
                &\in (1{\pm} \eps)\cdot \Pr_\mu[Y{=}1\mid (X,Z){=}(x,z)].%
            \end{align*}
            Steps 1 and 2 show that for each $t$, 
            \[
                \Pr\insquare{\evG}\geq 1-4\delta \sigma-2\delta\geq 1-6\delta.
            \]
            Fix any classifier $f\in \evF$, conditioned on $\evG$ the following inequalities hold 
            \begin{align*}
                &\abs{
                            \Ex_{\eta_w}\insquare{h(f(X,Z),Y,Z) \mid (X,Z)\in \expl_t }%
                            - 
                            \Ex_{\mu}\insquare{h(f(X,Z),Y,Z)\mid (X,Z)\in \expl_t }
                    }\\
                &\quad= \abs{\sum_{(x,z)\in \expl_t} \sum_{y\in \zo} h(f(x,z),y,z) \inparen{
                    \Pr_{\eta_w}[(X,Y,Z)=(x,y,z)] - \Pr_{\mu}[(X,Y,Z)=(x,y,z)]
                }} \\
                &\quad\leq \sum_{(x,z)\in \expl_t}\sum_{y\in \zo} \abs{
                        \Pr_{\eta_w}[(X,Y,Z)=(x,y,z)] - \Pr_{\mu}[(X,Y,Z)=(x,y,z)]
                    }\tag{Using that the range of $h$ is $[0,1]$ and triangle inequality}\\ 
                &\quad\leq \sum_{(x,z)\in \expl_t\setminus R}\sum_{y\in \zo} \abs{
                        \Pr_{\eta_w}[(X,Y,Z)=(x,y,z)] - \Pr_{\mu}[(X,Y,Z)=(x,y,z)]
                    }\\
                &\quad\quad +\sum_{(x,z)\in \expl_t\cap R}\sum_{y\in \zo} \abs{
                        \Pr_{\eta_w}[(X,Y,Z)=(x,y,z)] - \Pr_{\mu}[(X,Y,Z)=(x,y,z)]
                    }\\
                &\quad= \sum_{(x,z)\in \expl_t\setminus R}\sum_{y\in \zo} 
                    \abs{
                        \begin{array}{l}
                            \ \ \Pr_{S_t}[(X,Z)=(x,z)]\cdot \Pr_{(X,Y,Z)\sim L_0\cup\dots\cup L_t}\insquare{Y=y\mid (X,Z)=(x,z)} \\
                            - \ \ 
                            \Pr_{\mu}[(X,Y,Z)=(x,y,z)]
                        \end{array}
                    }\\
                &\quad\quad +\sum_{(x,z)\in \expl_t\cap R}\sum_{y\in \zo} \abs{
                        \Pr_{\eta_w}[(X,Y,Z)=(x,y,z)] - \Pr_{\mu}[(X,Y,Z)=(x,y,z)]
                    }\\
                &\quad\in \sum_{(x,z)\in \expl_t\setminus R}\sum_{y\in \zo} 
                    \abs{
                        \begin{array}{l}
                            \ \ (1\pm \eps)^2 \Pr_{\mu}[(X,Z)=(x,z)]\cdot \Pr_{(X,Y,Z)\sim \mu}\insquare{Y=y\mid (X,Z)=(x,z)} \\
                            - \ \ 
                            \Pr_{\mu}[(X,Y,Z)=(x,y,z)]
                        \end{array}
                    }\\
                &\quad\quad +\sum_{(x,z)\in \expl_t\cap R}\sum_{y\in \zo} \abs{
                        \Pr_{\eta_w}[(X,Y,Z)=(x,y,z)] - \Pr_{\mu}[(X,Y,Z)=(x,y,z)]
                    }
            \end{align*}
            \begin{align*}
                &\quad\leq \sum_{(x,z)\in \expl_t\setminus R}\sum_{y\in \zo} 
                    \abs{
                            (1\pm \eps)^2 - 1
                    }\cdot \Pr_{\mu}[(X,Y,Z)=(x,y,z)]\\ 
                &\quad\quad +\sum_{(x,z)\in \expl_t\cap R}\sum_{y\in \zo} \abs{
                        \Pr_{\eta_w}[(X,Y,Z)=(x,y,z)] - \Pr_{\mu}[(X,Y,Z)=(x,y,z)]
                    }\\
                &\quad\leq 3\eps\cdot \sum_{(x,z)\in \expl_t\setminus R}\sum_{y\in \zo} \Pr_{\mu}[(X,Y,Z)=(x,y,z)]\tag{Using that $(1-x)^2\leq 1-3x$ and $(1+x)^2\leq 1+3x$ for all $x\in [0,1]$}\\
                &\quad\quad +\sum_{(x,z)\in \expl_t\cap R}\sum_{y\in \zo} \abs{
                        \Pr_{\eta_w}[(X,Y,Z)=(x,y,z)] - \Pr_{\mu}[(X,Y,Z)=(x,y,z)]
                    }\\
                &\quad\leq 3\eps +\sum_{(x,z)\in \expl_t\cap R}\sum_{y\in \zo} \abs{
                        \Pr_{\eta_w}[(X,Y,Z)=(x,y,z)] - \Pr_{\mu}[(X,Y,Z)=(x,y,z)]
                    }\\ 
                &\quad\leq 3\eps +\sum_{(x,z)\in \expl_t\cap R}\sum_{y\in \zo} 
                        \inparen{{\Pr_{\eta_w}[(X,Y,Z)=(x,y,z)]} + {\Pr_{\mu}[(X,Y,Z)=(x,y,z)]}}\\
                &\quad\leq 3\eps + 2\eps + \sum_{(x,z)\in \expl_t\cap R}\sum_{y\in \zo} {\Pr_{\eta_w}[(X,Y,Z)=(x,y,z)]}.
                \yesnum\label{eq:lastEquation}
            \end{align*}
            It remains to upper bound $\sum_{(x,z)\in \expl_t\cap R}\sum_{y\in \zo} \abs{\Pr_{\eta_w}[(X,Y,Z)=(x,y,z)]}$.
            Towards this, observe that 
            \begin{align*}
                \sum_{(x,z)\in \expl_t\cap R}&\sum_{y\in \zo} \Pr_{\eta_w}[(X,Y,Z)=(x,y,z)]\\
                &\qquad = \sum_{(x,z)\in \expl_t\cap R}\sum_{y\in \zo}  \Pr_{S_t}[(X,Z)=(x,z)]\cdot \Pr_{(X,Y,Z)\sim L_0\cup L_1\cup\dots\cup L_t}\insquare{Y=y\mid (X,Z)=(x,z)}\\
                &\qquad \leq 2\sum_{(x,z)\in \expl_t\cap R}\Pr_{S_t}[(X,Z)=(x,z)]\\
                &\qquad = 2\Pr_{S_t}[(X,Z)\in \expl_t\cap R]\\
                &\qquad \leq 4\eps. \tag{Using \cref{eq:bound133}}
            \end{align*}
            Substituting this in \cref{eq:lastEquation}, implies that conditioned on $\evG$, for any $f\in \evF$
            \[
                \abs{
                        \begin{array}{l}
                            \Ex_{\eta_w}\insquare{h(f(X,Z),Y,Z) \mid (X,Z)\in \expl_t }
                            - 
                            \Ex_{\mu}\insquare{h(f(X,Z),Y,Z)\mid (X,Z)\in \expl_t }
                        \end{array}
                    }\leq 7\eps.
            \]
            Since $\Pr[\evG] \geq 1-6\delta$, the result follows by rescaling $\delta$ and $\eps$ by 6 and 7 respectively.

    \subsection{Proof of \cref{thm:fairness}}\label{sec:proofof:thm:fairness}
        In this section, we prove \cref{thm:fairness}.
        For ease of reference, we restate \cref{thm:fairness} below.

        \thmFairness*

        Recall that we assume that \cref{assump:f0_is_feasible} holds with constants $\alpha,\lambda\in (0,1]$.
        This implies that the classifier $f_0\in \cF$ satisfies the $\alpha$-FDR constraint and has a selection rate of at least $\lambda$, i.e., 
        \[
            \frac{\Pr_\mu\insquare{f_0(X,Z)\neq Y}}{\Pr_\mu\insquare{f_0(X,Z) = 1}}\leq \alpha
            \quad\text{and}\quad 
            \Pr_\mu\insquare{f_0(X,Z)=1}\geq \lambda.
            \yesnum\label{eq:prop_f02}
        \]
        Where, as in the rest of the proof, we use $\Pr_\mu[\cdot]$ to denote $\Pr_{(X,Y,Z)\sim \mu}[\cdot]$ and $\Ex_\mu[\cdot]$ to denote $\Ex_{(X,Y,Z)\sim \mu}[\cdot]$. 
        We also assume that the hypothesis class $\cF$ is derived from some base hypothesis class $\cB\subseteq \zo^{\cX}$ (\cref{sec:theoretical_results}).
        We need to show that for any utility function $\util$ and numbers $\alpha$, $\lambda$, $\delta,$ and $\tau$,
            if $n$ is sufficiently large, then at every iteration $t\in \inbrace{1,2,\dots}$
            with probability at least $1-\delta$,
        the classifier $f_t$ learned by the framework in \cref{alg:main_algorithm} satisfies 
        \[
            \util_{\mu,t}(f_t,z) \geq \max_{0\leq i\leq t-1} \util_{\mu,t}(f_i,z).
        \]
        Where $\util_{\mu,t}(f,z)$ is the utility of classifier $f$ over samples $(X,Y,Z)\sim \mu$ conditioned on the facts that $(X,Z)\in \textsc{Exploit}_t$ and $Z=z$.
        We will show that the following lower bound on $n$ is sufficient 
        \[
            n \geq \frac{12\abs{D}}{\tau \eps^3}\cdot \frac{1}{\alpha_{\rm explore} \lambda} \cdot \log{\frac{\abs{D}}{\delta\sigma}}.
            \yesnum\label{eq:lb_on_n2}
        \]
        Fix any two iterations $t_1>t_2$.
        We use $\eta(w,t)$ to denote the distribution $\eta_w$ in the $t$-th iteration.
        Since $\cF$ is derived from $\cB$, each hypothesis $f_t\in \cF$ is a tuple $(f_{t1},f_{t2},\dots,f_{tp})$ of hypothesis from $\cB$.
        Using this and the fact that the classifier $f_t$ learned by \cref{alg:main_algorithm} is an optimal solution of \prog{prog:main_algorithm}, it follows that 
        the corresponding hypothesis $f_{t,z}$, for each $z\in [p]$, is a solution to the following optimization program (which is an alternate version of \prog{prog:main_algorithm} specifically over samples in the $z$-th group):
        for each $z\in [p]$
        \begin{align*}
            f_{t,z} 
            ~~=~~ &\arg\max_{h\in \cF} \ \ \util_{\eta(w,t)}\inparen{f,\gamma,z},\yesnum\label{prog:group_wise_prog}\\
                     \quad \st,& \ 
                         \Pr_{\eta(w,t)}\insquare{h \neq  y\mid h=1, Z=z}\leq \alpha_{\rm exploit}+\eps
                    \quad\text{and}\quad
                        \Pr_{\eta_w}\insquare{h=1, Z=z}
                        \geq \lambda - \eps.
        \end{align*}
        Let $\evH$ be the event that for both $t\in \inbrace{t_1,t_2}$ and any $h\colon \zo\times \zo\times [p]\to [0,1]$
        \[
                \Pr\insquare{
                    \forall_{f\in \cF},\quad 
                    \abs{
                        \begin{array}{l}
                            \ \ \Ex_{\eta_w}\insquare{h(f(X,Z),Y,Z) \mid (X,Z)\in \expl_t }\\
                            - \ \ 
                            \Ex_{\mu}\insquare{h(f(X,Z),Y,Z)\mid (X,Z)\in \expl_t }
                        \end{array}
                    }
                    \leq \eps
                }
                \geq 1-\delta.
            \]
        By \cref{lem:conc_inequality} $\Pr\insquare{\evH}\geq 1-2\delta$.
        Conditioned on $\evH$ %
        \begin{align*}
            \util_\mu\inparen{f_{t_1},\gamma,z}
            & =  
            \sum_{i,j\in \zo} \gamma_{ij} \Pr_{\mu}\insquare{f_{t_1}(X,Z)=i\text{ and } Y=j\text{ and } Z=z}
            \tag{Using \cref{def:utility}}\\
            & \geq 
            \sum_{i,j\in \zo} \gamma_{ij}{\Pr_{\eta(w,t_1)}\insquare{f_{t_1}(X,Z){=}i\text{ and } Y{=}j\text{ and } Z{=}z} - \gamma_{ij}\eps}\tag{Using \cref{lem:conc_inequality}}\\ 
            & \geq  
            \sum_{i,j\in \zo} \gamma_{ij} \Pr_{\eta(w,t_1)}\insquare{f_{t_1}(X,Z)=i\text{ and } Y=j\text{ and } Z=z} - O(\eps) \tag{Using $\gamma_{ij}$ is a bounded constant}\\
            & =   
            \util_{\eta(w,t_1)}\inparen{f_{t_1},\gamma,z} - O(\eps) \tag{Using \cref{def:utility}}\\ %
            & =   
            \util_{\eta(w,t_1)}\inparen{f_{t_2},\gamma,z} - O(\eps). 
        \end{align*}
            Where the last equality holds because $f_{t_1}$ and $f_{t_2}$ are feasible for \prog{prog:group_wise_prog} at the $t_1$-th iteration and $f_{t_1}$ is the optimal solution of \prog{prog:group_wise_prog} at the $t_1$-th iteration.
            Proceeding with the above chain of inequalities, it follows that
        \begin{align*}
            \util_\mu\inparen{f_{t_1},\gamma,z}
            &\quad \geq  \quad 
            \util_{\eta(w,t_1)}\inparen{f_{t_2},\gamma,z} - O(\eps) \tag{Using \cref{def:utility}}\\
            &\quad =\quad  \sum_{i,j\in \zo} \gamma_{ij} \Pr_{\eta(w,t_1)}\insquare{f_{t_2}(X,Z)=i\text{ and } Y=j\text{ and } Z=z} - O(\eps)\\
            \qquad\qquad\quad~~~
            &\quad =\quad  \sum_{i,j\in \zo} \gamma_{ij} \inparen{\Pr_{\mu}\insquare{f_{t_2}(X,Z)=i\text{ and } Y=j\text{ and } Z=z} -\eps} - O(\eps)
            \tag{Using \cref{lem:conc_inequality}}\\
            &\quad =\quad  \sum_{i,j\in \zo} \gamma_{ij} \Pr_{\mu}\insquare{f_{t_2}(X,Z)=i\text{ and } Y=j\text{ and } Z=z} - O(\eps) \tag{Using that $\gamma_{ij}$ is a bounded constant}\\
            &\quad =\quad  \util_{\mu}\inparen{f_{t_1},\gamma,z} - O(\eps). \tag{Using \cref{def:utility}}
        \end{align*}
        The result follows as $\evH$ holds with probability at least $1-O(\delta)$.

    \subsection{Proof of \cref{thm:convergence_finite}}
    \label{sec:proofof:thm:convergence_finite}
        In this section, we prove \cref{thm:convergence_finite}.
        For ease of reference, we restate \cref{thm:convergence_finite} below.

        \thmConvergenceFinite*
        Since \cref{assump:f0_is_feasible} holds with constants $\alpha,\lambda\in (0,1]$, $f_0$ satisfies the $\alpha$-FDR constraint and has a selection rate of at least $\lambda$, i.e., 
        \[
            \frac{\Pr_\mu\insquare{f_0(X,Z)\neq Y}}{\Pr_\mu\insquare{f_0(X,Z) = 1}}\leq \alpha
            \quad\text{and}\quad 
            \Pr_\mu\insquare{f_0(X,Z)=1}\geq \lambda.
            \yesnum\label{eq:prop_f03}
        \]
        Where we use $\Pr_\mu[\cdot]$ to denote $\Pr_{(X,Y,Z)\sim \mu}[\cdot]$ and $\Ex_\mu[\cdot]$ to denote $\Ex_{(X,Y,Z)\sim \mu}[\cdot]$. 
        To prove \cref{thm:convergence_finite}, given $\eps,\delta,\tau\in (0,1]$ and a function $g\colon D\times \cF \to \R_{\geq 0}$, 
        we need to show that if $n$ and $t$ are sufficiently large (\cref{eq:params_proof3}), 
        then with probability at least $1-\delta$, 
        the classifier $f_t$ learned by the data collection and prediction framework in \cref{alg:main_algorithm}, satisfies the following inequality:
        \[
            \forall_{z\in [p]},\quad 
            \util_\mu(f_t,\gamma,z) 
            \geq 
            \util_\mu(\optOffline{}^{(\alpha+\alpha')},\gamma,z) - \eps.
        \]
        Where $\optOffline^{(\alpha+\alpha')}$ is the optimal offline classifier defined in Equation~\ref{eq:optOffline}.
        \cref{thm:convergence_finite} claims that the following lower bounds on $t$ and $n$ are sufficient.
        \[
            t\geq \frac{1}{\sigma(z)}%
            \quad\text{and}\quad 
            n\geq \frac{12\abs{D}}{\tau \eps^3}\cdot \frac{1}{\alpha_{\rm explore}^2 \lambda} \cdot \log{\frac{\abs{D}}{\delta\sigma}}.
            \yesnum\label{eq:params_proof3}
        \]
        The bounds on $t$ and $n$ serve different purposes. 
        The lower bound on $t$ ensures that $\expr_t$ has no samples from the $z$-th group.
        The lower bound on $n$ ensures that the classifier $f_t$, trained over the samples in $\expl_t$ has a high utility on samples from the $z$-th group drawn from the underlying distribution $\mu$.

        Fix any index $z\in [p]$.
        We first prove the first claim. 
        Concretely, we will prove that with probability at least $1-\delta$, $\expr_t$ does not contain any samples $(x,z)$ satisfying 
        $\mu(x,z)\geq \frac{\eps}{\abs{D}}$.
        Fix any sample $(x,z)$ from the $z$-th group in $\expr_0$.
        In each iteration $t$, where $(x,z)\in \expr{}_t$, its weight $w(x,z)$ increases by at least $\sigma(z)$ as the marginal distribution of $L_t$ has density at least $\sigma(z)$ on $(x,z)$.
        Hence, $(x,z)$ can be in the exploration region for at most $\frac{1}{\sigma(z)}$ iterations.
        Thus, after $t\geq \frac{1}{\sigma(z)}$ iterations all items from the $z$-th group are in the exploitation region.

        Now, we can bound the utility of $f_t$ using \cref{lem:conc_inequality}.
        Let $\evH$ be the event in \cref{lem:conc_inequality}, conditioned on $\evH$, we have the following lower bounds
        \begin{align*}
            \util_\mu\inparen{f_t,\gamma,z}
            &= 
            \sum_{i,j\in \zo} \gamma_{ij} \Pr_{\mu}\insquare{f_t(X,Z)=i\text{ and } Y=j\text{ and } Z=z}\tag{Using \cref{def:utility}}\\
            &\geq  
        \sum_{i,j\in \zo} \gamma_{ij} \inparen{\Pr_{\eta_w}\insquare{f_t(X,Z){=}i\text{ and } Y=j\text{ and } Z{=}z} {-} \eps}\tag{Using \cref{lem:conc_inequality}}\\
            &=  
            \sum_{i,j\in \zo} \gamma_{ij} \Pr_{\eta_w}\insquare{f_t(X,Z)=i\text{ and } Y=j\text{ and } Z=z} - O(\eps) \tag{Using $\gamma_{ij}$ is a bounded constant}\\
            &=  
            \util_{\eta_w}\inparen{f_t,\gamma,z} - O(\eps)\tag{Using \cref{def:utility}}\\ %
            &=  
            \util_{\eta_w}\inparen{\optOffline^{(\alpha+\beta)},\gamma,z} - O(\eps).
        \end{align*}
            Where the last equality is implied by the facts that $\optOffline^{(\alpha+\beta)}$ is feasible for \prog{prog:main_algorithm}, $f_t$ is the optimal solution of \prog{prog:main_algorithm}, and $\util_{\eta_w}\inparen{\cdot,\gamma,z}$ is the objective function of \prog{prog:main_algorithm}.
            Proceeding with the above chain of inequalities, we get the following lower bound.
        \begin{align*}
            \util_\mu\inparen{f_t,\gamma,z}
            &\geq   
            \sum_{i,j\in \zo} \gamma_{ij} \inparen{\Pr_{\eta_w}\insquare{\optOffline^{(\alpha+\beta)}(X,Z)=i\text{ and } Y=j\text{ and } Z=z} - \eps} - O(\eps) 
                \tag{Using \cref{lem:conc_inequality} and \cref{def:utility}}\\
            &\geq   
            \sum_{i,j\in \zo} \gamma_{ij} \Pr_{\mu}\insquare{\optOffline^{(\alpha+\beta)}(X,Z)=i\text{ and } Y=j\text{ and } Z=z} - O(\eps)\tag{Using $\gamma_{ij}$ is a bounded constant}\\
            & = 
            \util_{\mu}\inparen{\optOffline^{(\alpha+\beta)},\gamma,z} - O(\eps).
            \tag{Using \cref{def:utility}}
        \end{align*}
        The result now follows as $\evH$ holds with probability at least $1-O(\delta)$.

\vspace{-2mm}

\section{Other related work} \label{sec:other_related_work}
    Approaches from related fields like active learning and fair classification can be employed in the partial feedback setting, but suffer from some drawbacks that we discuss in this section.

    \vspace{-1mm}
    
    \smallskip\noindent\textbf{Fairness classification.}
    To reduce performance disparity across demographic groups, fair classification \cite{zafar2015fairness,zafar2017fair,AgarwalBD0W18,celis2019classification} and multi-armed bandit \cite{celis2019personalization,vishakha2021fairness} approaches can also potentially be used.
    However, fairness constraints are informative of outcome bias only when evaluated over datasets whose empirical distribution is close to the underlying population distribution.
    Since partial feedback results in distribution shifts, the above assumption may not hold, and learning using only fairness constraints would not guarantee low disparity and high utility \cite{kallus2018residual}.
    Our framework allows for supplementing fairness constraints with additional exploration to address this issue.

    \vspace{-1mm}

    \smallskip\noindent\textbf{Active learning.}    
    Active learning-based data collection approaches assume that labels can be queried for each sample at a \textit{fixed cost} \cite{william2022adaptive,arteaga2022moredata,abernethy2022active,shen2022metricfair,anahideh2022fairactivelearning,sharaf2022promoting}. 
    These approaches, once again, focus on achieving high long-term utility by addressing gaps in the available outcome (empirical) distribution. 
    The main difference between our work and prior active learning approaches is the presence of the FDR constraint which ensures that iteration-wise utility is also continuously high despite exploration. 
    Additionally, unlike the standard setup for active learning, our work assumes that labeling costs depend on the outcome and the classifier. For instance, a false positive can be significantly more costly to the institution than a false negative (as in the case of loan defaults).
    The adaptive exploration approach of \citet{yangadaptive} falls within the category of active learning and \cref{sec:empirical} demonstrates the drawbacks of this approach - empirical analysis shows that the algorithm of \citet{yangadaptive} achieves significantly lower cumulative utility and higher statistical rate disparity than our framework.

    \vspace{-1mm}

    \smallskip\noindent\textbf{Classification using selective labels.}
    For judicial bail decisions, \citet{kleinberg2018human} train a model using past judges' bail decisions with true outcomes only available for released defendants.
    Unlike our work, they do not incorporate outcome information from positively classified samples into classifier training. 
    \citet{arteaga2018selective} suggest a similar approach for data augmentation 
    that exploits the human decision-makers' decisions in regions where they are accurate and trains a classifier for regions where they are not. 
    Their approach also has additional drawbacks: it assumes that all decision-makers have similar behavior, which is not true in our setting where the decision-makers are classifiers that can be different across different iterations.
    Extrapolation \cite{coston2021selective} or reweighting \cite{li2020inferring} methods for partial feedback settings have also been proposed to
    impute the labels for unlabeled samples.
    Imputation of this form, however, assumes that all samples are appropriately represented in the labeled data distribution.
    This assumption does not hold in settings where historical biases have led to the under-representation of certain groups in the labeled data.
    {Algorithms for ensuring fairness in sequential decision-making tasks -- modeled as Markov decision processes (MDPs) -- are also relevant in this setting \cite{wen2021MDP}. 
    While these may be used for decision-making, their practical use may be limited by the assumption that all involved individuals' actions follow a known MDP.}

    \vspace{-1mm}

    \smallskip\noindent\textbf{Data augmentation.}    
    Beyond the data collection approaches discussed in \cref{sec:intro}, certain related methods often turn towards other sources to obtain information about outcome data of under-explored populations.
    This includes methods of data collection using human annotators \cite{arteaga2022moredata} or augmenting available data using third-party signals \cite{fong2022AUC}; however, these approaches are often impractical as they can only provide proxies for the true outcomes, which themselves can encode social biases \cite{srinivasan2021biases}.
        One recent approach also studies how to distribute a specified data-collection budget among different data sources
        \cite{feiyang2024DataSelection}. 
        In this case, the data collection takes place \textit{before} the prediction phase 
        and does not rely on the predictions being made. 
        The main challenge of partial feedback setting, however, is that data collection and learning are necessarily intertwined.
        For instance, in the credit-lending application, banks require data on past loan repayment rates which themselves are determined by the predictions made by the bank (i.e., the loans it gives out).
        Prior approaches that do not account for this causal relation between prediction and data collection are bound to be either ineffective in improving prediction \mbox{performance and/or wasteful in data collection.}

\section{Discussion and Conclusion}\label{sec:discussion}
    We provide a framework for data collection and learning that obtains high prediction utility in every iteration and gathers outcome data for previously unobserved samples.
    Our framework ensures that false positive errors are bounded and employs fairness mechanisms to improve the exploration of under-represented groups.
    The explicit focus on exploration is crucial since the under-represented subpopulations are usually not random subsets of the domain, but rather are composed of individuals who have historically been denied equal opportunities.
    In applications where classification is employed to streamline the decision-making process, explicit focus on exploration, along with learning using available training data, can lead to improved prediction accuracy and fairness.     
    We next discuss certain practical advantages, additional exploration strategies, distribution and outcome shifts, multi-class classification, and limitations of our framework.
    
    \smallskip\noindent
    \textbf{Short-term gains and long-term utility.}
    {
    An important advantage of our framework is that it uses both exploitation 
    and exploration 
    strategies in every iteration.
    Using available data to make accurate predictions over the exploit regions, we ensure that the utility 
    is high in every iteration.
    And by randomly selecting samples in the explore region, we gather data that improves prediction utility in later iterations.
    We also provide exploration strategies that can lead to higher utility over the explore region by using the available classifier (e.g. $g=g^{\rm fair}$ leads to better utility over the Adult dataset).
    With appropriate choices, our framework shows gains in both short-term and long-term utilities.}
    
    \smallskip\noindent\textbf{Fairness.} 
    {
    As discussed earlier, fairness can be incorporated 
    by training classifier $f_t$ to satisfy fairness constraints or by using exploration strategies that encode fairness notions.
    In practice, a combination of both could be the most effective way of tackling biases.
    }
    Fairness constraints when learning $f_t$ can also encode an implicit form of exploration, since
    they may encourage increased selection of minority individuals to satisfy the constraints.
    However, as noted in prior work \cite{KallusMZ20}, implicit exploration using fairness constraints may not be sufficient for collecting outcome data about minority groups.
    In Section~\ref{sec:empirical}, we observe that complementing fairness constraints in exploitation with fairness strategies in exploration is the most effective solution to reduce disparities.
    Hence, standard fair classification performance can be improved by using additional exploration.

    \smallskip\noindent\textbf{Outcome observations.} Our framework assumes that for samples predicted positively at any iteration $t$, the true outcomes will be observed before time $t + 1$.
    However, in real-world applications, outcomes for different samples might be observed at different time intervals.
    For instance, loan default is defined as whether a loan was repaid or not within a certain number of years. 
    However, different people could repay at different times. 
    We make the assumption that all outcomes are observed before the next iteration for ease of analysis.
    Moreover, we also assume that the data observed is accurate whereas, in practice, data inevitably has recording errors that can have adverse effects \cite{ChenKMSU19,wang2020robust,celis2021adversarial,keswani2023strategic}.
    Future work can further explore methods to model settings where samples are observed at different time periods and possibly contains errors. %
    
    \smallskip\noindent\textbf{Other fair exploration strategies.} 
    We discussed a few exploration strategies in Section~\ref{sec:algorithm} that either assume no information (uniform sampling), strategies that use the classifier ($g^{\rm clf}(x,z) \propto \beta + (1-\beta) f_t(x,z)$), and fair exploration strategies that use the classifier ($g^{\rm fair}(x,z) \propto (\beta+(1-\beta)f_t(x,z)) \cdot \Pr_\mu\insquare{Z=z \mid (X,Z)\in \expr_t}$).
    There are many other possible choices of exploration strategies and we discuss a few other fair strategies here.

    If the classifier has very low accuracy or large biases over the $\expr$ region, it would not make sense to use it in the exploration function.
    In such cases, if we still want to sample minority group individuals at a higher rate, then using the uniform sampling function would sample groups proportional to their representation in the $\expr_t$ region. 
    Hence, if these groups are under-represented in the $\expl_t$ region but over-represented in the $\expr_t$ region, they would be selected at a higher rate.
    Alternately, if the underlying population is demographically imbalanced and the institution wants to select an equal number of individuals from all groups for exploration, then it can use the exploration function $g(x,z) \propto  1/\Pr_\mu\insquare{Z=z \mid (X,Z)\in S_t}$.
    If the classifier does have reasonable accuracy for some $\expr_t$ samples and the institution wants to select an almost equal number of samples from all groups, then it can use the exploration function $g(x,z) \propto (\beta + (1-\beta) f_t(x,z)) \cdot 1/\Pr_\mu\insquare{Z=z \mid (X,Z)\in S_t}$.
    
    Depending on the context and application and question, different choices for $g$ would be relevant and useful. 
    Nevertheless, in all settings, it is important to select a function that {takes positive values} and ensure all demographic groups are appropriately represented among the explored samples.

    \smallskip\noindent\textbf{Multi-class classification}. We have mainly considered binary classification settings so far.
    However, our proposed algorithm can be used for multi-class classification as well.
    Suppose the set of all class labels is $\mathcal{Y}$ and outcomes are observed only if the prediction belongs to a subset $\mathcal{Y}' \subset \mathcal{Y}$. 
    In this case, at iteration $t$, Step~3 of \cref{alg:main_algorithm} can learn a multi-class classifier over the labeled weighted data. The remaining steps, i.e., determining the EXPLORE and EXPLOIT partitions and exploration remains unchanged. This is because computing weights for each sample (Step 6 in Algorithm~\ref{alg:main_algorithm}) would once again involve determining whether at least $\tau$ fraction of previous classifiers would have predicted a label in set $\mathcal{Y}'$ - if yes, this sample is put in EXPLOIT region, otherwise put it in EXPLORE region.
    With these simple modifications, Algorithm~\ref{alg:main_algorithm} can be used for multi-class classification settings.

    \smallskip\noindent\textbf{Choice of hypothesis class.} The choice of hypothesis class for \cref{alg:main_algorithm} is important. It encodes context-specific knowledge, such as, whether individuals in different groups have similar or different distributions of features and labels.
        On the one hand, if the distribution of features and labels is similar across groups, then we may want classifiers to not use protected attributes for predictions. 
        On the other hand, if the distribution of features and labels is different across groups, then we may want the classifiers to use the protected attributes for prediction.
        Depending on the application and available data, this choice can be made appropriately.

    \smallskip\noindent\textbf{Limitations of empirical analysis.} In our empirical analysis, we assume that the applicants arriving at every iteration are sampled i.i.d.\ from the dataset distribution.
    However, the distributions for the Adult and German datasets are likely to be different than the true population distributions.
    Since we do not have access to the true distribution and only have access to the given dataset, we are only able to simulate the setting where the initial data available to the framework $L_0$ is different than the dataset distribution (i.e., treating dataset distribution as true distribution).
    Future simulations on settings where true underlying distribution is available will be useful to assess the complete impact of exploration.
 
    \smallskip\noindent\textbf{Broader impact.} Our framework provides a method to collect outcome information about relatively unexplored subpopulations so that classifiers trained using observed data are accurate over the entire population and not just for subpopulations for whom data is available.
    It is important to note that fairness is primarily ensured for predefined protected attributes.
    Our framework might not ensure performance parity for groups that are not explicitly defined as ``protected'' and, hence, it would be crucial to first identify all group attributes for which observation disparities exist.
    
    Secondly, exploration in settings like credit lending comes with risks. 
    Providing a loan to a person who cannot pay it back can be harmful to the person's financial status and severely affect their credit score.
    We try to minimize this possibility by incorporating the decision of classifiers learned on exploit region during exploration, but the stochasticity of exploration can still make this scenario possible.
    While exploration stochasticity can be beneficial to the institution in terms of collecting data about the entire population, it would also be useful to scout additional exploration strategies that minimize risk for individuals and this can be a fruitful direction for future work on this topic.

\section*{Acknowledgements}
    This project is supported in part by NSF Award IIS-2045951.

\newpage
\bibliographystyle{ACM-Reference-Format} 
\bibliography{reference.bib}

\clearpage
\appendix

\section{Additional Discussion on Sample Complexity}\label{sec:additionalDiscussion}
    
        Our theoretical results (Theorems~\ref{thm:feasibility}, \ref{thm:fairness}, \ref{thm:convergence_finite}) do not make assumptions on the underlying distribution $\mu$, as additional assumptions may reduce the practical usefulness of the framework.
        This results in linear dependence on domain size $\abs{D}$, which can be large.
        The difficulty in proving such bounds
        is that they rely on training data being sampled i.i.d.\ from the underlying distribution $\mu$.
        However, in any non-trivial data collection framework, the specific samples collected are bound to depend on the observations made in the previous iterations.
        This dependence on past observations violates the independence assumption.
        
        That said, if the distribution $\mu$ is known to satisfy additional properties, then the number of samples required by the theoretical results may be reduced. 
        For instance, suppose the underlying distribution is ``smooth,'' as assumed by generalization guarantees of clustering algorithms (see Chapter 19 of \cite{shalev2014understanding}).
        In this case, one can first cluster the samples in the domain $D$ to obtain a set of clusters $C$ such that almost all samples in each cluster have the same label, and then use our framework with $C$ as the underlying domain (see Theorem 19.3 \cite{shalev2014understanding}).
        In this case, the dependence on $\abs{D}$ improves to $\abs{C}$, which is always smaller and the ratio $\frac{\abs{C}}{\abs{D}}$ depends on the desired accuracy and the ``smoothness'' of $\mu$.

\section{Implementation details}\label{sec:implementation_details}

        In this section, we provide additional details on how to implement Step~3 of \cref{alg:main_algorithm}.
        Step~3 involves learning a classifier that satisfies the FDR constraints and the fairness constraints if required.
        The classifier optimizes a given utility measure, and we show to implement this program if the utility measure corresponds to accuracy or revenue first.

        For accuracy, one can simply minimize a weighted logistic regression model over $L_t$, with FDR and fairness constraints.
        To implement this in practice (and for simulations in Section~\ref{sec:empirical}), we use Python's SLSQP program.
        Alternately, any common constrained optimization techniques can be employed here.
        
        To learn a classifier that maximizes revenue, we first optimize a constrained optimization program to obtain a model that accurately assigns likelihoods to all individuals and then adjust the classifier threshold over these likelihoods to maximize revenue.
        We can accomplish this process in the following manner:
        Firs partition $L_t$ into two equal random parts: $L_{t,1}$ and $L_{t,2}$.
        For learning likelihoods, we can again use constrained logistic regression, i.e.,
         first learn the parameters $\omega \in \R^d$ of the logistic model ($d$ is the number of features) which minimizes the weighted log-loss associated with $\omega$ over $L_{t,1}$, subject to the likelihoods satisfying the given $\alpha$-FDR constraint.
        The constrained optimization program can again be implemented using Python's SLSQP function.
        Then, using $L_{t,2}$, choose a likelihood threshold in $[0,1]$ (such that points with likelihood greater than the threshold are to be classified positively) which maximizes revenue$_{c_1, c_2}$.
        This method essentially corresponds to choosing an appropriate threshold from the model's ROC curve and has been used in other papers that optimize revenue for lending settings \cite{gramespacher2021employing}.

        \section{Additional Details and Results for the Adult and German Dataset} \label{sec:empirical_other}

        In this section, we provide the implementation details and additional results for the Adult and German datasets which were omitted from Section~\ref{sec:empirical}.

        \medskip\noindent
        \textbf{Description and pre-processing of Adult and German dataset.}
            For the Adult dataset, each individual is characterized by the following features: age, class of worker, educational attainment, marital status, occupation, place of birth, usual hours worked per week past 12 months, gender, and race.
            We pre-process the dataset to ensure that it is limited to the subset with only individuals belonging to the races white/Caucasian and black/African-American.
            All features, other than the protected attribute are also scaled to have a mean of 0 and a standard deviation of 1.

        For the German Credit dataset, the features are every individual's credit amount, duration, installment rate in percentage of disposable income, residential status, age, number of existing credits, number of people liable for, and gender.
        Once again, all features, other than the protected attribute are also scaled to have a mean of 0 and a standard deviation of 1.

        \medskip\noindent
        \textbf{Additional parameter details for our algorithm.}
        We implement the constrained optimization program using the method described in \cref{sec:implementation_details}.
        For the SLSQP function (used to solve the optimization program), we use  parameters ftol$=1e^{-3}$ and eps$=1e^{-3}$.

        \medskip\noindent
        \textbf{Implementation of $g^{\rm clf}$ and $g^{\rm fair}$ exploration strategies in \cref{alg:main_algorithm}.}
        As described in \cref{sec:adult}, the exploration function $g$ is either $g^{\rm clf}(x,z) \propto \beta + (1-\beta) f_t(x,z)$ or $g^{\rm fair}(x,z) \propto (\beta+(1-\beta)f_t(x,z)) \cdot \Pr_\mu\insquare{Z=z \mid (X,Z)\in \expr_t}$, depending on whether exploration fairness is used or not.
        Both functions require a choice of $\beta$ which ensures that every sample is assigned a non-zero exploration probability.
        We implement our approaches with $\beta = 0$, but instead of using binary outcome $f_t(x,z)$ in the above functions, we use the likelihood derived from the logistic model (described in \cref{sec:implementation_details}).
        Since the likelihood assigned by the logistic regression model is non-zero for every point, we ensure that $g$ takes positive values
        and simultaneously use classifier performance to assign exploration probabilities.

        \medskip\noindent
        \textbf{Implementation details for baselines.}
        We implement the baselines to correspond to our iterative setting and we mention the implementation details below.

        \begin{itemize}
            \item \textsc{opt-offline.} As mentioned earlier, the \textsc{opt-offline} baseline is trained using the initial part of the split dataset, i.e., $S_0$.
            This baseline simply maximizes revenue subject to the $\alpha$-FDR constraint and is implemented using Python's SLSQP function.
            \item \textsc{Fair-clf.} The \textsc{Fair-clf} baseline is trained to test an only exploitation algorithm with statistical parity constraints.
            It maximizes revenue subject to the $\alpha$-FDR constraint and statistical parity constraint and is trained over the available labeled dataset every iteration.
            This baseline is also implemented using Python's SLSQP function.
            \item \textsc{Kilbertus et al.} The algorithm from the \citet{kilbertus2020fair} paper is implemented using the stochastic batch gradient descent method, as suggested in their paper.
            We also use the demographic parity regularizer employed in their paper.
            As they recommend, we also use a logistic regression model for the classifier in this algorithm.
            The parameter $c$ in their algorithm is set to be 0.6 (similar to their experiments), the batch size is kept to 128, the learning rate is 0.01, number of iterations for gradient descent is also kept to 128.
            At every iteration, the parameters of the logistic model are updated using stochastic gradient descent over $B$ elements randomly sampled from the recently labeled elements.
            \item \textsc{Yang et al.} The algorithm from the \citet{yangadaptive} paper is implemented the code provided by the authors \footnote{\url{https://github.com/Yifankevin/adaptive_data_debiasing}}. 
            We had to make certain modifications to the code to make it suitable for our setting and we list the modifications below.
            The full algorithm presented in Appendix C of their paper uses a while loop over the entire dataset.
            However, since the entire dataset is not available in our setting, we modify this step to a for loop over the iterations.
            In each iteration $t$, we provide their algorithm with the batch of samples $S_t$ that arrive at the beginning of iteration $t$.
            All other components of their code are kept unchanged.
            \item \textsc{Rateike et al.} The algorithm from the \citet{rateike2022don} paper is implemented the code provided by the authors \footnote{\url{https://github.com/ayanmaj92/fairall}}. 
            We execute this baseline using the initial data and batch size configuration specified in \cref{sec:empirical}. All other components are kept unchanged and we use the same parameters as specified in the original code.
        \end{itemize}

\begin{table}
    \centering
    \small
    \caption{Comparison of the overall performance of all methods on the Adult dataset with gender as the protected attribute. The average revenue per iteration across all repetitions, average FDR, and average acceptance rate disparity (statistical rate) are reported along with the standard deviation of all metrics in parenthesis.}
    \medskip 
    \begin{tabular}{lcccc}
    \toprule
    & \multicolumn{4}{c}{Protected attribute - Gender} \\    
    Method & \specialcell{Revenue\\(in thousands)} & FDR  & Stat. Rate  & \specialcell{TPR\\Disparity} \\
    \midrule
Algorithm~\ref{alg:main_algorithm}- no fairness constraint & 75.6 (15.0)  & .15 (.02)  & .09 (.02)  & .15 (.04) \\ 
Algorithm~\ref{alg:main_algorithm}- only exploit fairness & 78.1 (12.4)  & .15 (.02)  & .07 (.02)  & .07 (.03) \\ 
Algorithm~\ref{alg:main_algorithm}- only explore fairness & 77.9 (12.5)  & .15 (.02)  & .08 (.02)  & .11 (.05) \\ 
Algorithm~\ref{alg:main_algorithm}- both fairness constraints &  78.0 (13.0)  & .15 (.02)  & .07 (.02)  & .08 (.03)  \\ 
\midrule
Baseline - \textsc{Opt-offline}  & 80.4 (9.0)  & .15 (.02)  & .10 (.02)  & .15 (.06)  \\ 
Baseline - \textsc{Fair-clf}  &  72.4 (9.4)  & .13 (.02)  & .02 (.01)  & .03 (.02)\\ 
\textsc{Kilbertus et al.}  & 66.1 (12.1)  & .20 (.03)  & .15 (.02)  & .23 (.04) \\  
\textsc{Yang et al.}   & -45.3 (11.6)  & .47 (.08)  & .09 (.02)  & .02 (.01) \\ \textsc{Rateike et al.}   & -17.1 (7.4)  & .12 (.01) &  .02 (.01)  & .02 (.01) \\       
    \bottomrule
    \end{tabular} 
    \label{tab:overall_adult_gender}
\end{table}

\begin{figure}
    \centering
    \includegraphics[width=\linewidth]{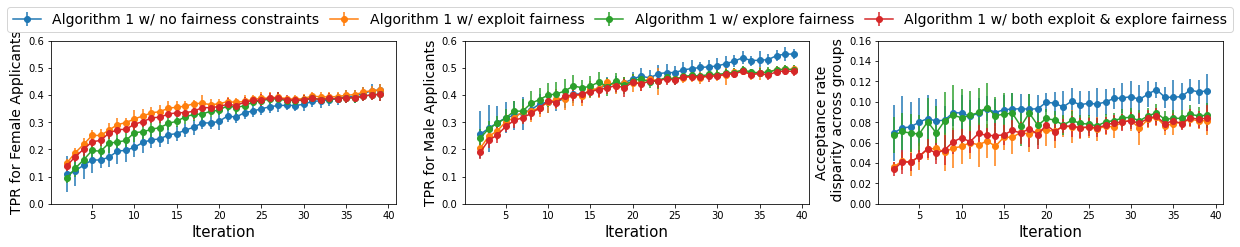}
    \caption{Iteration-wise performance of all variants of Algorithm~\ref{alg:main_algorithm} (with or without each of explore and exploit fairness constraints) on the Adult dataset with gender as the protected attribute. All algorithms can be seen to improve TPR for both groups.}
    \label{fig:results_explore_exploit_gender_only_ours}
\end{figure}

        \medskip\noindent
        \textbf{Additional results.}
        \textit{Performance comparison over the Adult dataset with gender as the protected attribute.}
        Overall performance on the Adult dataset for gender protected attribute is presented in Table~\ref{tab:overall_adult_gender} and iteration-wise performance is presented in Figure~\ref{fig:results_explore_exploit_gender_only_ours}.
        
        For gender, \cref{alg:main_algorithm} with exploit fairness achieves the lowest average statistical rate and true positive rate disparity and highest cumulative revenue among all variants.
        \cref{alg:main_algorithm} with both explore and exploit fairness also has similar performance.
        Hence, for both protected attributes, using both fairness constraints lead to high revenue while ensuring small performance disparities. 
        
        Figure~\ref{fig:results_explore_exploit_gender_only_ours} 
        present the iteration-wise performance of our algorithms.
        The first two plots in the figure show that 
         TPR increases with increasing iterations for all demographic groups.
        TPR is also generally larger in the initial iterations when using exploit fairness, but is overtaken by or is similar to the TPR of \cref{alg:main_algorithm} with no fairness constraints in the final iterations.
        Hence, fairness constraints can assist in accelerated data collection in the initial iterations, but once sufficient data is available, it seems to have a similar TPR as the variant with no fairness constraints.
        The third plot in Figure~\ref{fig:results_explore_exploit_gender_only_ours} also show that using exploit fairness results in the smallest statistical rate in all iterations.

        \textit{Iteration-wise comparison of Algorithm~\ref{alg:main_algorithm} to baselines.}
        We present plots for iteration-wise comparison of all methods and baselines with respect to the following metrics: cumulative revenue, FDR, TPR disparity, group-wise TPR, and statistical rate.
        We exclude the \textsc{Yang et al} and \textsc{Rateike et al} baselines from these plots as it achieves negative revenue and assigns positive prediction to a large fraction of samples, which makes it difficult to analyze the performance of other algorithms through the plots.

\begin{figure*}
    \centering
    \includegraphics[width=\linewidth]{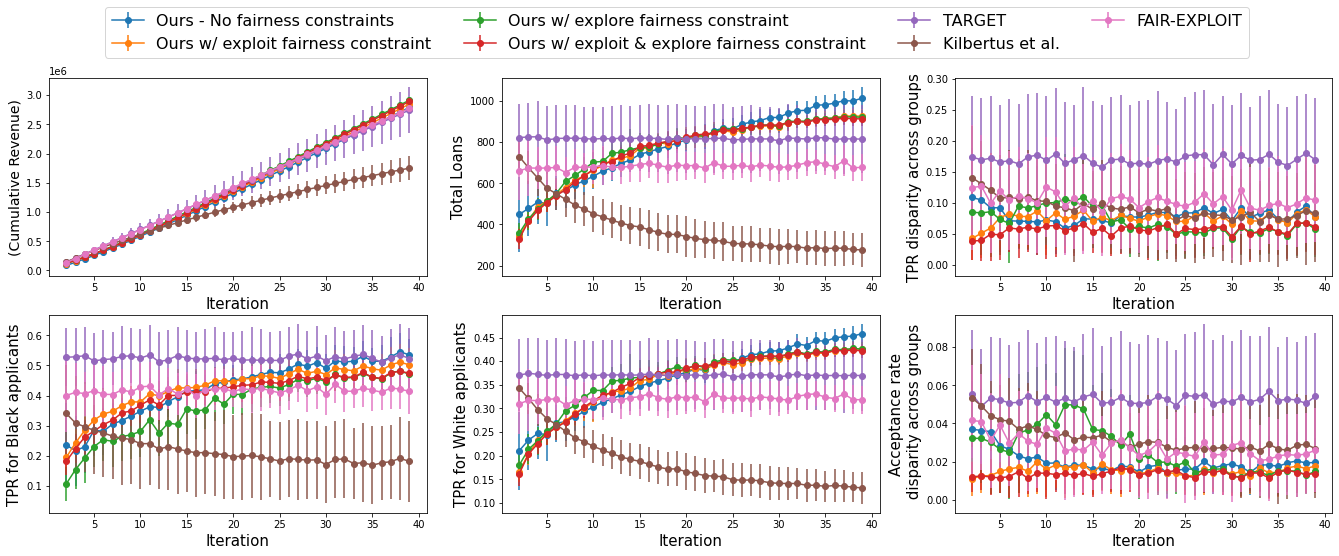}
    \caption{Performance of all versions of Algorithm~\ref{alg:main_algorithm} (with or without each of explore and exploit fairness constraints) and baselines on the Adult dataset with race as the protected attribute.}
    \label{fig:results_explore_exploit_race_all}
\end{figure*}

\begin{figure*}
    \centering
    \includegraphics[width=\linewidth]{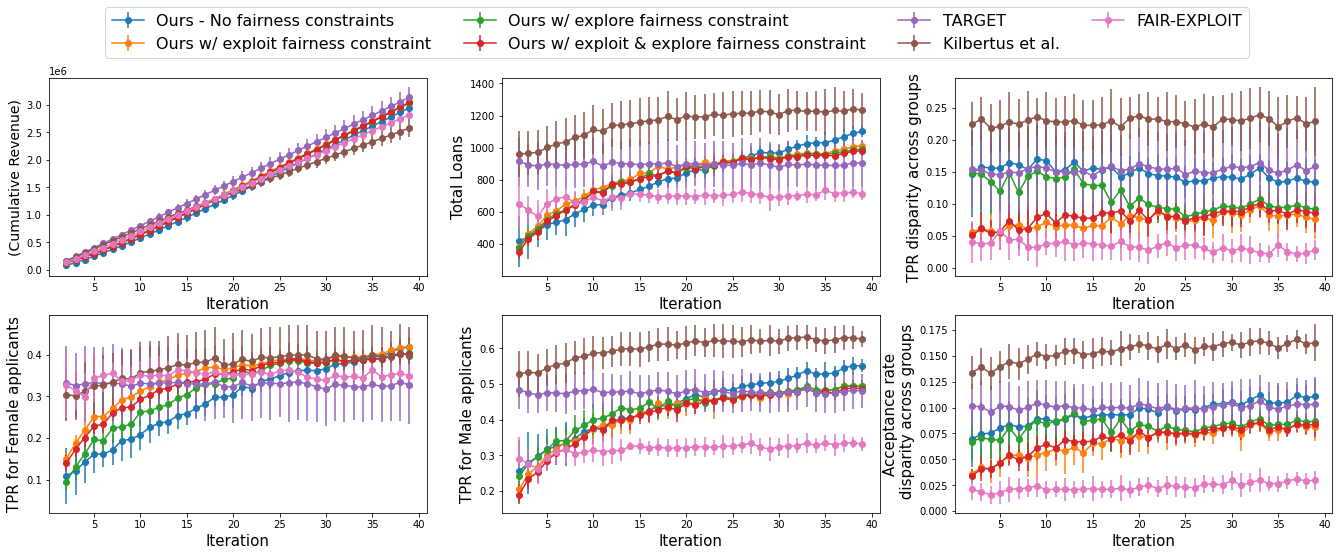}
    \caption{Performance of all versions of Algorithm~\ref{alg:main_algorithm} (with or without each of explore and exploit fairness constraints) and baselines on the Adult dataset with gender as the protected attribute.}
    \label{fig:results_explore_exploit_gender_all}
\end{figure*}

\begin{figure*}
    \centering
    \includegraphics[width=\linewidth]{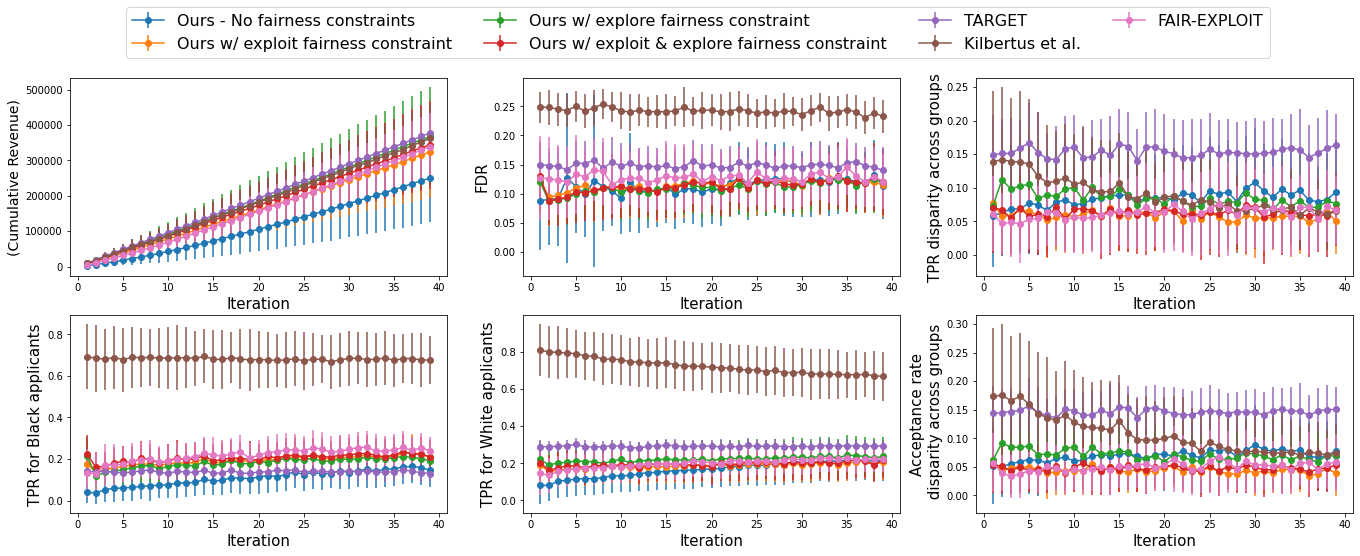}
    \caption{Performance of all versions of Algorithm~\ref{alg:main_algorithm} (with or without each of explore and exploit fairness constraints) and baselines on the German dataset with gender as the protected attribute.}
    \label{fig:results_explore_exploit_german_all}
\end{figure*}

        Figure~\ref{fig:results_explore_exploit_race_all} presents the performance of all methods over the Adult dataset with race as the protected attribute.
        Figure~\ref{fig:results_explore_exploit_gender_all} presents the performance over the Adult dataset with gender as the protected attribute.
        Figure~\ref{fig:results_explore_exploit_german_all} presents the performance over the German dataset.
        All sets of plots show that the cumulative revenue from our methods are similar to the \textsc{opt-offline} baseline across all iterations.
        Furthermore, unlike other methods, the TPR of our method consistently improves with increasing iterations.
        Baselines \textsc{fair-clf} and \textsc{kilbertus et al} achieve high TPR in certain cases but in many settings, their TPR stagnates or decreases with increasing iterations.

\begin{figure*}
    \centering
    \includegraphics[width=\linewidth]{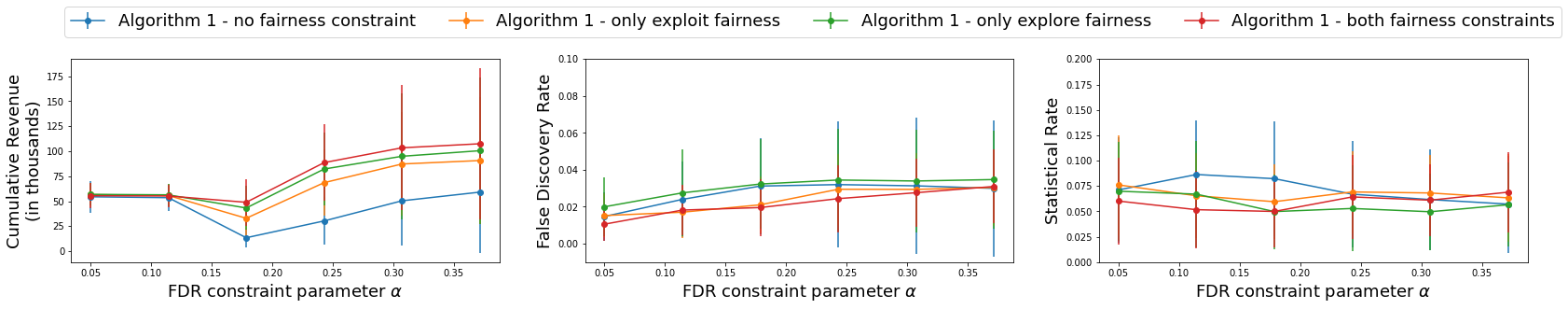}
    \caption{Performance of all versions of Algorithm~\ref{alg:main_algorithm} (with or without each of explore and exploit fairness constraints) with respect to different $\alpha$ parameters on the Adult dataset with race as the protected attribute.}
    \label{fig:results_adult_diff_fdr_race}
\end{figure*}

\begin{figure*}
    \centering
    \includegraphics[width=\linewidth]{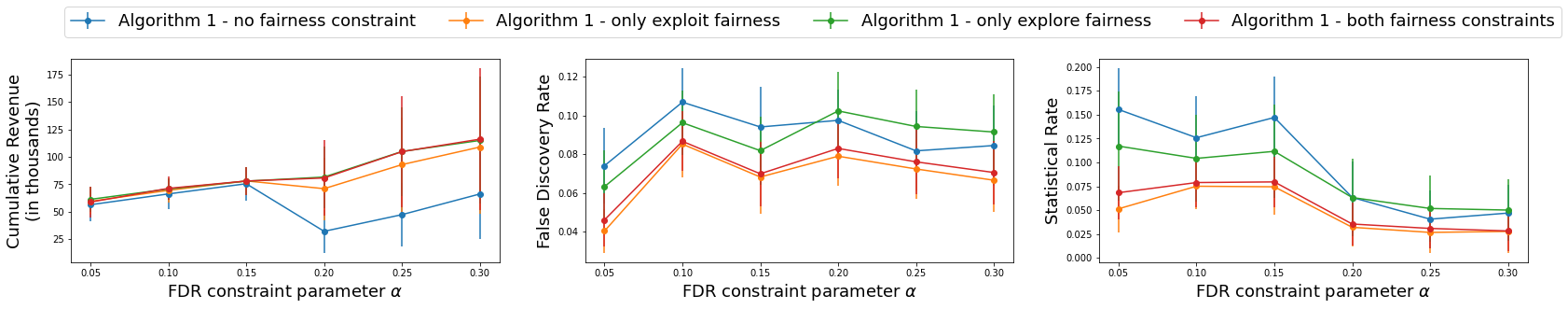}
    \caption{Performance of all versions of Algorithm~\ref{alg:main_algorithm} (with or without each of explore and exploit fairness constraints) with respect to different $\alpha$ parameters on the Adult dataset with gender as the protected attribute.}
    \label{fig:results_adult_diff_fdr_gender}
\end{figure*}

        \noindent\textit{Variation in performance of Algorithm~\ref{alg:main_algorithm} with $\alpha$.}
        The results presented in \cref{sec:empirical} used $\alpha=0.15$ for the $\alpha$-FDR constraint.
        We also present performance variation with respect to $\alpha$ over the Adult dataset.
        Figures~\ref{fig:results_adult_diff_fdr_race} and \ref{fig:results_adult_diff_fdr_gender} present the results for race and gender protected attributes respectively.
        As $\alpha$ increases, the framework is allowed to make more false positive errors which result in larger variability in revenue. However, across all iterations, our methods with appropriate fairness constraints result in low performance disparity across protected attribute groups.

\end{document}